%% file: main.tex
\documentclass[letterpaper,journal]{IEEEtran}
\usepackage{amsmath,amsfonts}
\usepackage{algorithmic}
\usepackage{algorithm}
\usepackage{array}
\usepackage[caption=false,font=normalsize,labelfont=sf,textfont=sf]{subfig}
\usepackage{textcomp}
\usepackage{stfloats}
\usepackage{url}
\usepackage{verbatim}
\usepackage{graphicx}
\usepackage{cite}
\usepackage{booktabs}
\usepackage{multirow}
\usepackage{siunitx}
\usepackage{xcolor}
\usepackage{hyperref}

\DeclareSIUnit\flops{FLOPs} % Floating point operations per second
\DeclareSIUnit\macs{MACs} % Multiply-Accumulate operations per second
\definecolor{tumblue}{RGB}{0,101,189}
\definecolor{tumgreen}{RGB}{162,173,0}
\definecolor{tumorange}{RGB}{227,114,34}

\usepackage{tikz}
\usepackage{textcomp}
\usepackage{lipsum}
\newcommand\copyrighttext{%
	\footnotesize \textcopyright~2024 IEEE.  Personal use of this material is permitted.  Permission from IEEE must be obtained for all other uses, in any current or future media, including reprinting/republishing this material for advertising or promotional purposes, creating new collective works, for resale or redistribution to servers or lists, or reuse of any copyrighted component of this work in other works.
}
\newcommand\copyrightnotice{%
	\begin{tikzpicture}[remember picture,overlay]
		\node[anchor=south,yshift=4pt] at (current page.south) {\fbox{\parbox{\dimexpr\textwidth-\fboxsep-\fboxrule\relax}{\copyrighttext}}};
	\end{tikzpicture}%
}

\begin{document}

\title{DPFT: Dual Perspective Fusion Transformer for Camera-Radar-based Object Detection}

\author{
    Felix Fent, \and
    Andras Palffy, \and
    Holger Caesar
    \thanks{Felix Fent is with the Technical University of Munich, School of Engineering and Design, Institute of Automotive Technology and Munich Institute of Robotics and Machine Intelligence (MIRMI), Germany. (\href{mailto:felix.fent@tum.de}{felix.fent@tum.de})}% 
    \thanks{Andras Palffy is with Perciv AI and Delft University of Technology, Microwave Sensing, Signals and Systems (MS3) Group, Department of Microelectronics, The Netherlands}% 
    \thanks{Holger Caesar is with Delft University of Technology, 
    Intelligent Vehicles (IV) Section, Department of Cognitive Robotics, The Netherlands}% 
}

% The paper headers
\markboth{IEEE TRANSACTIONS ON INTELLIGENT VEHICLES}%
{Fent \MakeLowercase{\textit{et al.}}: Dual Perspective Fusion Transformer for Camera-Radar-based Object Detection}

\maketitle

\begin{abstract}
The perception of autonomous vehicles has to be efficient, robust, and cost-effective. However, cameras are not robust against severe weather conditions, lidar sensors are expensive, and the performance of radar-based perception is still inferior to the others. Camera-radar fusion methods have been proposed to address this issue, but these are constrained by the typical sparsity of radar point clouds and often designed for radars without elevation information. We propose a novel camera-radar fusion approach called Dual Perspective Fusion Transformer (DPFT), designed to overcome these limitations. Our method leverages lower-level radar data (the radar cube) instead of the processed point clouds to preserve as much information as possible and employs projections in both the camera and ground planes to effectively use radars with elevation information and simplify the fusion with camera data. As a result, DPFT has demonstrated state-of-the-art performance on the K-Radar dataset while showing remarkable robustness against adverse weather conditions and maintaining a low inference time. The code is made available as open-source software under \href{https://github.com/TUMFTM/DPFT}{https://github.com/TUMFTM/DPFT}.
\end{abstract}

\begin{IEEEkeywords}
Perception, \and Object Detection, \and Sensor Fusion, \and Radar, \and Camera, \and Autonomous Driving
\end{IEEEkeywords}

\copyrightnotice

\section{Introduction}
\label{sec:intro}
\IEEEPARstart{A}{utonomous} driving is a promising technology that has the potential to increase safety on public roads and provide mobility to people for whom it was previously not accessible. However, leveraging this technology requires autonomous vehicles to operate safely within a multitude of different environmental conditions. These conditions include everyday driving situations such as nighttime driving or driving under severe weather conditions, but also critical situations where the autonomous vehicle (AV) has to react quickly or maintain general functionality after a sensor failure.

The perception of most autonomous driving systems is based on either camera or light detection and ranging (lidar) sensors. While camera sensors are cost-effective, they depend on ambient light and do not provide depth information~\cite{Arnold2019}. In contrast, lidar sensors provide accurate measurements of the surroundings but come at a high cost. More importantly, neither camera nor lidar sensors are robust against severe weather conditions like rain, fog, or snow~\cite{Yoneda2019}. On the other hand, radio detection and ranging (radar) sensors are cost-effective and robust against challenging environmental conditions but do not yet provide comparable object detection qualities as lidar or camera-based perception methods due to their low spatial resolution and high noise level~\cite{LiuY2023}.

A potential solution to overcome the limitations of individual sensor technologies is the combination of multiple sensor modalities, also referred to as sensor fusion. Nevertheless, sensor fusion remains challenging due to inherent differences between the camera and radar sensors, such as the perceived dimensionality (2D vs. 3D), data representation (point cloud vs. grid), and sensor resolution~\cite{Yao2023}. \IEEEpubidadjcol

\begin{figure}[tb]
    \centering
    \includegraphics[width=8.85cm]{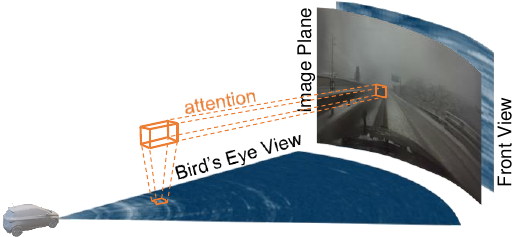}
    \caption{Illustration of the dual perspective fusion procedure. The 4D radar cube is projected onto a front and bird's eye view to create a parallel and perpendicular perspective to the camera image. This simplifies the camera-radar fusion and maintains the complementary sensor features. Object features are queried from these perspectives via an attention mechanism and used to regress 3D detections.}
    \label{fig:intro}
\end{figure}

In this paper, we propose a novel camera and radar sensors fusion method to provide a robust, performant, yet cost-effective method for 3D object detection. While camera-radar fusion has been done before~\cite{Yao2023}, previous methods mostly rely on radar point cloud data, thus suffering from a sparse data representation and facing the challenge of combining images with point clouds. On the other hand, fusion approaches that utilize raw radar data solely rely on radar data in a bird's eye view (BEV) representation. Therefore, they are fusing data from the image plane with data from a perpendicular BEV plane on one side and discarding the advantages of modern 4D radar sensors on the other. 

Our proposed method overcomes these limitations by fusing camera data with raw radar cube data to mitigate the differences in sensor resolution and benefit from a structured grid representation for both sensor modalities. However, directly consuming the raw radar cube would be unfeasible due to its high demand for computational resources. Therefore, we developed a projection method that reduces the 4D radar cube to two 2D grids while maintaining important features and providing a low sensitivity to input noise. As a result, the proposed fusion architecture utilizes radar data from both a BEV and a front-view perspective as shown in Figure~\ref{fig:intro}. With this dual perspective approach, we create a corresponding data source to the image plane to support camera-radar fusion and incorporate data from the BEV plane to exploit all radar dimensions. All three data inputs are then fed through a ResNet feature extractor and subsequent Feature Pyramid Network (FPN) neck before they are combined in the fusion module. However, our method does not require a combined feature space but queries 3D objects directly from these individual perspectives, thus preventing the loss of information caused by a uniform feature space or raw data~fusion~\cite{Liu2023}. To enable this, we introduce a modified deformable attention~\cite{Zhu2020} mechanism that allows both cartesian and spherical reference point projection to realize a modality-agnostic sensor fusion.

In summary, our main contributions are three-fold:

\begin{itemize}
    \renewcommand\labelitemi{\textbullet}
    \item We propose an efficient sensor fusion approach that projects the radar cube onto two perspectives, thus simplifying the camera-radar fusion, avoiding the limitations of sparse radar point clouds, and leveraging the advantages of 4D radar sensors.
    \item We are the first to fuse 4D radar cube data with image data by proposing a novel fusion method that does not rely on a common BEV representation to fuse camera and radar data.
    \item Experiments show that our method achieves state-of-the-art results in severe weather conditions on the challenging K-Radar dataset thus offering greater robustness and lower inference times than previous methods.
\end{itemize}

\section{Related Work}
\label{sec:sota}
The proposed method combines the complementary features of camera and radar sensors to create a robust, performant, and cost-effective method for 3D object detection. However, to understand the motivation behind the proposed Dual Perspective Fusion Transformer (DPFT), it is important to understand the concepts and limitations of unimodal object detection methods and available datasets.

\subsection{Camera-Radar Datasets}
While there are many datasets within the autonomous driving domain, most of them do not include radar sensor data~\cite{Yao2023}. The nuScenes~\cite{Caesar2020} dataset only provides 3D radar point clouds and has been criticized for its limited radar data quality~\cite{Engels2021, Schumann2021}. The RadarScenes~\cite{Schumann2021} dataset provides higher-quality radar data but only on a point cloud level and does not provide object annotations. Both the View-of-Delft~\cite{Palffy2022} as well as the TJ4DRadSet~\cite{Zheng2022} datasets provide 4D radar data and corresponding bounding boxes but do not include raw radar data. The CARRADA~\cite{Ouaknine2021}, RADIATE~\cite{Sheeny2021}, and CRUW~\cite{Wang2021.2} datasets are one of the few proving cube-level radar data but are limited to 3D radar data and do not provide 3D object annotations. The RADIal~\cite{Rebut2022} dataset provides raw 4D radar data but originally only included 2D bounding box annotations. Even if 3D annotations were recently added by Liu~et~al.~\cite{LiuY2023}, the RADIal dataset does not support the retrieval of 4D radar cube data, has a limited extent, and does not include data within severe weather conditions, which is one of the main motivations for radar applications. For these reasons, the K-Radar~\cite{Paek2022} dataset is the only suitable dataset for our experiments. The dataset itself includes raw (cube-level) radar data from a 4D radar sensor as well as the data from two lidar sensors, 4 stereo cameras, one GNSS, and two IMU units. In addition, it provides 3D annotated bounding boxes for 34994 frames sampled from 58 different driving scenes and is split into \SI{49.9}{\percent} train and \SI{50.1}{\percent} test data.

\subsection{Camera-based 3D Object Detection}
Camera-based monocular 3D object detection methods can be divided into three major categories: data lifting, feature lifting, and result lifting methods~\cite{Ma2023, Qian2022}.

Data lifting methods directly lift 2D camera data into 3D space to detect objects within it~\cite{Ma2023}. Out of those, pseudo-lidar methods~\cite{Wang2019} are most commonly used to transform camera images into 3D point clouds. Besides that, learning-based approaches~\cite{Srivastava2019} can be used for data lifting, and even most feature lifting methods~\cite{Philion2020, Harley2023} can directly be applied to image data.

Feature lifting methods first extract 2D image features, which are then lifted into 3D space to serve as the basis for the prediction of 3D objects~\cite{Ma2023}. Within this category, there are two dominant lifting strategies: one "pushes" (splatting) the features from 2D into 3D space~\cite{Philion2020} and the other "pulls" (sampling) the 3D features from the 2D space~\cite{Harley2023}.

Result lifting methods are characterized by the fact that they first estimate the properties of the objects in the 2D image plane and then lift the 2D detections into 3D space~\cite{Ma2023}. Inspired by the taxonomy used within the field of 2D object detection, these methods can be further divided into one-stage and two-stage detectors. One-stage detectors regress 3D objects directly from 2D image features and are typically characterized by fast inference speeds. Representative methods of this category are anchor-based detectors~\cite{Brazil2019} or anchor-free models like~\cite{Zhou2019}. Two-stage detectors first generate region proposals before they refine those proposals to predict 3D objects~\cite{Ma2023}. Methods from this category can use either geometric priors~\cite{Chen2015, Chen2016} or model-based priors~\cite{Naiden2019}.

Even if different strategies have been developed over the years, the biggest challenge for camera-based 3D object detection remains the lifting from 2D to 3D space due to the inability of camera sensors to directly measure depth information~\cite{Arnold2019}. Furthermore, camera sensors are susceptible to illumination changes and severe weather conditions~\cite{Arnold2019}, limiting their robustness in field applications. 

\subsection{Radar-based 3D Object Detection}
Radar sensors, in contrast to cameras, are robust against severe weather conditions~\cite{Yoneda2019} and are able to measure not only depth information but also intensities and relative velocities via the Doppler effect. This is due to the fact that radar sensors perceive their environment by actively emitting radio wave signals and analyzing their responses~\cite{Yao2023}. However, this analysis requires multiple processing steps, which is why radar-based 3D object detection methods are categorized by the data level they are operating on~\cite{LiuY2023}. The first category of methods operates directly on the raw analog-to-digital converted (ADC) radio wave signals. These ADC signals are then converted from the temporal to the spatial domain using a Discrete Fast Fourier Transformation (DFFT). The resulting data representation is a discrete but dense radar cube and the basis for the second type of detection methods. Finally, this data can be further reduced by only considering data points with high response values, leading to a spare point cloud representation and the input to the third (and most common) type of methods~\cite{Yao2023}.

Methods operating on the raw ADC signals are rare due to limited data availability, high memory requirements, and the abstract data format. Even if Yang~et~al.~\cite{Yang2023} achieved promising results on the RADIal~\cite{Rebut2022} dataset, Liu~et~al.~\cite{LiuY2023} showed that ADC data has no advantages over radar cube data. Thus, the benefits of replacing the DFFT with a neural network remain questionable.

Detection methods utilizing cube-level radar data can be subdivided into those using 2D, 3D, or 4D radar data. Methods utilizing 2D radar data use either range-azimuth (RA)~\cite{Wang2021, Dong2020, Kaul2020} or range-doppler (RD)~\cite{Ng2020, Decourt2022} measurements, while 3D methods either use multiple 2D projections~\cite{Major2019, Gao2021} or the whole range-azimuth-doppler (RAD) cube~\cite{Palffy2020, Zhang2021}. However, none of the above mentioned methods are used for 3D, but only 2D object detection, and neither of those utilizes the elevation information of modern 4D (3+1D) radar sensors.

Methods relying on radar point clouds are the most common type of detectors and can be further divided into grid, graph, and point-based methods. Grid-based methods~\cite{Tan2023, Dreher2020, Khler2023, LiuJ2023} discretize the point cloud space to derive a regular grid from the sparse point cloud. Graph-based methods~\cite{Ulrich2022, Svenningsson2021, Fent2023} create connections (edges) between the points (vertices) to utilize graph neural networks (GNNs) for object detection tasks. Lastly, point-based methods~\cite{Tilly2020, Scheiner2020, Dubey2022, Danzer2019, Schumann2020} use specialized network architectures to directly detect objects within the sparse irregular radar point clouds.

Generally, radar-based object detection methods are robust against severe weather conditions but do not yet achieve competitive performance values. This is mainly due to the radar's lower spatial resolution, higher noise level, and limited capability to capture semantic information.

\subsection{Camera-Radar Fusion for 3D Object Detection}
The complementary sensor characteristics of camera and radar sensors make them promising candidates for sensor fusion applications. These fusion methods can be divided into data-level, object-level, and feature-level fusion methods.

Data-level fusion aims to directly combine the raw data from both sensor modalities. Following this approach, Nobis~et~al.~\cite{Nobis2019} was the first to propose a camera-radar fusion model that projected the radar points into the camera image and used a hierarchical fusion strategy to regress objects from it. On the other side, Bansal~et~al.~\cite{Bansal2022} projected the semantic information of the camera image onto the radar point cloud (similar to PointPainting~\cite{Vora2020}) and detected objects within the enriched radar data. Nevertheless, data-level fusion is associated with a high loss of information due to the differences in sensor resolution~\cite{Liu2023} and challenging due to different data representations and dimensionalities.

Object-level fusion addresses these challenges by using two separate networks for both modalities independently and only combining their detection outputs. Using this technique, Jha~et~al.~\cite{Jha2019} fused 2D objects from a camera and radar branch, while Dong~et~al.~\cite{Dong2021} combined an object-level with a data-level fusion approach to detect 3D objects on a proprietary dataset. Most recently, Zhang~et~al.~\cite{Zhang2024} fused the outputs of a radar-based method~\cite{Zhang2021} with the detections of a camera-lidar fusion method and achieved state-of-the-art results on the K-Radar dataset~\cite{Paek2022}. However, their method, solely relying on camera and lidar data,  outperformed the radar fusion method only slightly, thus showing that the capabilities of object-level fusion are limited. This is because object-level fusion exclusively depends on the final detection outputs, neglecting any intermediate features~\cite{Yao2023}. As a result, the final detection quality relies heavily on the performance of the individual modules and does not fully utilize complementary sensor features~\cite{Yao2023}.

Feature-level fusion aims to combine the advantages of both methods by first extracting features from each modality separately, fusing them at an intermediate level, and finally predicting objects based on their combined feature space. Therefore, it allows to address individual sensor aspects and benefits from a combination of their unique properties. However, finding a suitable feature space to combine both modalities remains challenging. Besides early attempts to combine region proposals from camera and radar branches~\cite{Nabati2021, Cui2021, Kim2020} or feature-level fusion on the image plane~\cite{Chang2020, Yadav2020}, most recent methods focus on a bird's eye view (BEV) feature~representation.

Using a BEV feature representation, Harley~et~al.~\cite{Harley2023} proposed a method to combine rasterized ("voxelized") radar point cloud data with camera data and outperformed their camera baseline on the nuScenes~\cite{Caesar2020} dataset. Similarly, Zhou~et~al.~\cite{Zhou2023} fused rasterized and temporally encoded radar point cloud data with image data in the BEV space and reported an increased detection quality. However, both methods utilize only 3D radar data, not considering modern 4D radar sensors. Addressing this issue, both Xiong~et~al.~\cite{Xiong2023} as well as Zheng~et~al.~\cite{Zheng2023} proposed a method to fuse camera and 4D radar point cloud data in a BEV space. While achieving good results on the TJ4DRadSet~\cite{Zheng2022} and View-of-Delft~\cite{Palffy2022} dataset, these methods solely rely on radar point cloud data. However, radar point cloud data is not only difficult to fuse due to its irregular, sparse data structure but also contains significantly less information, which is lost during signal processing and adverse to accurate environment perception~\cite{lim2019}.

To prevent this loss of information, Liu~et~al.~\cite{LiuY2023} proposed a method to fuse raw radar data with camera image data, similar to our approach. However, their method relies on an intermediate BEV representation, which increases the demand on computational resources and limits their ability to encode various 3D structures~\cite{Huang2023}. Moreover, their method does not utilize the elevation information of modern 4D radar sensors, but solely relies on radar data in the range-azimuth (BEV) plane. To overcome these limitations, we propose a novel method that does not require a uniform feature representation and exploits all radar dimensions.

\section{Methodology}
\label{sec:method}
\begin{figure*}[t]
  \centering
  \includegraphics[width=18.13cm]{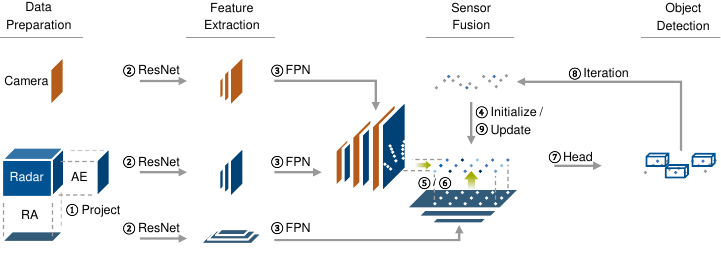}
  \caption{The DPFT model overview shows the essential steps to fuse camera data with raw 4D radar data and retrieve objects from it. First \raisebox{.5pt}{\textcircled{\raisebox{-.8pt}{1}}}, the data of the 4D radar cube is projected onto the range-azimuth (RA) and azimuth-elevation (AE) plane. Second \raisebox{.5pt}{\textcircled{\raisebox{-.8pt}{2}}}, the two radar perspectives and the camera data are fed through individual ResNet backbones to extract essential features from them. In the \raisebox{.5pt} {\textcircled{\raisebox{-.8pt}{3}}} step, Feature Pyramid Networks (FPN) are used to align the dimensions of the multi-level feature maps. To fuse the features of the different perspectives, a set of query points is initialized in 3D space in the \raisebox{.5pt}{\textcircled{\raisebox{-.8pt}{4}}} step and projected onto the different perspectives in the \raisebox{.5pt}{\textcircled{\raisebox{-.8pt}{5}}} step. After that, the features hit by the projection points are fused in the associated query points, using deformable attention \raisebox{.5pt}{\textcircled{\raisebox{-.8pt}{6}}}. A classification and regression head is used in \raisebox{.5pt}{\textcircled{\raisebox{-.8pt}{7}}} to retrieve bounding boxes from the queried features. Finally, the regressed bounding box positions are used as new query points in step {\textcircled{\raisebox{-.8pt}{8}}} and their features are updated {\textcircled{\raisebox{-.8pt}{9}}} in an iterative process to refine the bounding box proposals.}
  \label{fig:model}
\end{figure*}

The Dual Perspective Fusion Transformer (DPFT) is designed to address the main challenges of multimodal sensor fusion, which are caused by the differences in the perceived dimensionality, data representations, and sensor resolutions. First, it utilizes raw cube-level radar data to preserve as much information as possible and lower the resolution differences between camera and radar data. Second, cube-level radar data is given in a structured grid representation, thus avoiding the fusion of point cloud and image data. Third, two projections are created from the 4D radar cube. One parallel to the image plane to support the fusion between camera and radar and another perpendicular to it to preserve the complementary radar information. Besides that, the model design aims to achieve a low inference time and is designed with no interdependencies between the two modalities such that the overall model remains operational even if one sensor modality fails. However, to achieve that, multiple steps are required, which are shown in Figure~\ref{fig:model} and explained in the following.

\subsection{Data Preparation}
The input data itself poses the greatest challenge for multimodal sensor fusion due to the differences in data resolution and dimensionality. Camera sensors capture the environment as a projection onto the 2D image plane, while radar sensors typically capture measurements in the range-azimuth (BEV) plane. Broadly speaking, these two perception planes are perpendicular to one another, which makes them difficult to fuse due to their small intersection. To counteract this, our method is built on 4D radar data with three spatial dimensions and one Doppler dimension. This allows us to create a physical relationship between the two data sources. However, working with 4D data is not ideal for two reasons. First, lifting camera data into 3D space is challenging due to the missing depth information, and second, processing high dimensional data has a high demand on computational resources. Resolving this dilemma, the radar data is projected onto the range-azimuth plane as well as the azimuth-elevation plane. This way, we can create a complementary data source to the camera data while reducing the data size and creating a physical relationship between the image and the BEV plane to regress 3D objects.

To address the challenges associated with diverging data formats and sensor resolutions, our method is based on raw (cube-level) radar data. Usually, radar data is given as an irregular, sparse point cloud with a few hundred points per sample, while camera data is represented in a structured grid format with millions of pixels. Not only is it difficult to fuse these two data formats, but a fusion is also associated with a high loss of information or computational overhead~\cite{Liu2023}. Furthermore, radar point clouds are the results of a multistage signal processing chain (explained in Section~\ref{sec:sota}) during which a lot of information is lost and which deteriorates perception performance~\cite{LiuY2023}. Therefore, our method utilizes raw (cube-level) radar data, avoiding the loss of information, creating a uniform data representation, and lowering the differences in data resolution.

Following this idea, the 4D radar cube is projected onto the range-azimuth (RA) and azimuth-elevation (AE) plane. However, to avoid the loss of important information and minimize the sensitivity to input noise, the design of the projection (dimensional reduction) follows a three-step process. First, a set of 30 initial radar features was defined that were proven to be significant to radar-based perception by previous studies~\cite{Schumann2017, Scheiner2018}. Secondly, a model was trained on all 30 radar features before the weights of the first model layer were analyzed to determine the importance of individual features to the converged model. Lastly, a sensitivity analysis was conducted where noise was added to individual input features and the changes in the output were monitored to determine the sensitivity of the model to input noise. As a result, the maximum, median, and variance of the amplitude and Doppler values were chosen to be extracted during the radar data projection. In addition, the first and last three cells of the radar cube are cut off to avoid DFFT artifacts in the AE projection. Besides that, the image data is rescaled to an input height of 512 pixels using bilinear interpolation to lower the demand on computational resources.

\subsection{Feature Extraction}
The multimodal input data is fed to consecutive backbone and neck models to deduce expressive features for the desired detection task. Every input is fed to an individual backbone model resulting in three parallel backbones. The purpose of the backbone networks is the extraction of expressive, higher-dimensional features for the subsequent sensor fusion and is chosen to be a ResNet~\cite{He2016} architecture. Since the standard ResNet implementation resizes the inputs to a height of 256, the resulting feature maps of all inputs have similar spatial dimensions. In addition, multi-scale feature maps are extracted from intermediate backbone layers (to detect objects at different scales) and skip connections are used to directly pass the input data to the neck models~\cite{Liu2019}. More specifically, a ResNet-101 is used for the camera data and a ResNet-50 for both radar data inputs. The larger image backbone is chosen because of the higher image data resolution compared to the radar data. All backbones have been pre-trained on the ImageNet database~\cite{Deng2009} and a single 1x1 convolution layer is added in front of the radar backbones to make them compatible with the six feature dimensions of the radar data.

The neck models are responsible for feature alignment and ensuring homogeneous feature dimensions. They align the feature dimensions of the multi-scale feature maps and the sensor raw data, which is required for the subsequent sensor fusion. In addition, it also exchanges information between the four feature maps (from three backbone models and the raw input data). For this purpose, a Feature Pyramid Network~\cite{Lin2017} with an output feature dimension of 16 is used.

\subsection{Sensor Fusion}
Our sensor fusion model allows the direct querying of fused features from the individual inputs and the retrieval of objects from them. Therefore, a combined intermediate feature space is not required. To achieve this, multi-head deformable attention~\cite{Zhu2020} is used, which was originally developed for camera-based object detection. This method projects reference points onto camera images to query features from the surrounding pixels. Therefore, this method allows to attend to a fixed number of keys in an image (or feature map), regardless of their spatial size. While this projection was originally designed for a pinhole camera model, we introduce a spherical reference point projection to utilize it for low-level radar data.

Our resulting fusion module consists of five distinct steps. First, the reference points are initialized as a set of 400 evenly distributed 3D points in a polar space with feature values sampled from a uniform distribution and cover the entire field of view (FoV) of the sensor. Next, the reference points are fed to a self-attention layer to allow the exchange of information between queries, which becomes important during the iterative refinement. After that, the reference points are projected onto the camera and the dual radar perspectives in the third step. Based on these projections, deformable cross-attention is used to query features from the (positional encoded) multi-level feature maps. In the last step, the queried features are passed through a feed-forward network (FFN) before they are combined in a max pooling layer. Besides that, each of the attention and FFN layers includes dropout, addition, and normalization layers. With this approach, multiple sensors from different modalities can be fused as long as a projection of the query points onto the sensor feature maps exists.

\subsection{Object Detection}
The detection head predicts object bounding boxes based on the fused query features and is separated from the fusion module to allow for multi-task applications. Following~\cite{Wang2022, Chen2023, Zhu2020}, we use an interactive output refinement process where the predicted bounding box centers and the previous query features are used for another three attention cycles. As a result, we get object bounding boxes represented by their 3D center point $(x, y, z)$, size $(l, w, h)$, heading angle $\theta$, and class label. The detection head design follows the example of other sparse detectors~\cite{Wang2022, Chen2023} and consists of three consecutive linear layers. However, DPFT uses a specific activation function for each bounding box component. The center point prediction utilizes an identity function due to its unrestricted value range, the bounding box size uses a ReLu~\cite{Hahnloser2000} activation function, and the heading angle is predicted by a hyperbolic tangent function. This is due to the fact that the heading angle is not predicted directly but rather split into its $\sin{\theta}$ and $\cos{\theta}$ components, since it is shown that the model training benefits from a continuous output space~\cite{ZhouY2019}. The class label is predicted by a sigmoid activation function and chosen to be the maximum across all classes.

\subsection{Model Training}
The model training uses a set-to-set loss with a one-to-one matching as introduced by DETR~\cite{Carion2020}. The loss function itself is composed of a focal loss~\cite{Lin2020} for classification and an L1 regression loss for all bounding box components~\cite{Wang2022}. The loss weights for these two terms are set to one such that the final loss function can be written as:

\begin{equation}
  \mathcal{L} = \mathcal{L}_{\text{class}} + \mathcal{L}_{\text{box}}.
  \label{eq:loss}
\end{equation}

The optimization scheme uses an AdamW~\cite{Loshchilov2017} optimizer with a learning rate of \SI{1e-4} and a constant learning rate throughout the training. All models are trained with a batch size of 4 and a maximum of 200 epochs ($\sim$\SI{72}{\hour}).

\section{Results}
\label{sec:results}
All reported results are achieved on the K-Radar~\cite{Paek2022} test set and are in line with the official evaluation scheme, which is based on the KITTI~\cite{Geiger2012} protocol. For comparability, the benchmark results of Table~\ref{tab:benchmark} were obtained on the original version (revision v1.0) of the dataset, while all other results were achieved on the revised version (revision v2.0) of the dataset, which is preferred since it includes corrected object heights and previously missing object labels. For development purposes, we split the train data into \SI{80}{\percent} train and \SI{20}{\percent} validation data, the test set remains unmodified.

\begin{table*}[t]
  \caption{3D object detection results for the test data of the K-Radar dataset revision v1.0.}
  \label{tab:benchmark}
  \centering
  \begin{tabular}{@{}p{0.15\textwidth}p{0.08\textwidth}p{0.07\textwidth}p{0.07\textwidth}p{0.07\textwidth}p{0.07\textwidth}p{0.07\textwidth}p{0.07\textwidth}p{0.07\textwidth}p{0.07\textwidth}@{}}
    \toprule
    \multirow{2}{*}{Method} & \multirow{2}{*}{Modality} & \multicolumn{1}{l}{\multirow{2}{*}{Norm.}} & \multirow{2}{*}{Overcast} & \multicolumn{1}{l}{\multirow{2}{*}{Fog}} & \multicolumn{1}{l}{\multirow{2}{*}{Rain}} & \multicolumn{1}{l}{\multirow{2}{*}{Sleet}} & \multicolumn{1}{l}{\multirow{2}{*}{\begin{tabular}[c]{@{}l@{}}Light\\ Snow\end{tabular}}} & \multirow{2}{*}{\begin{tabular}[c]{@{}l@{}}Heavy\\ Snow\end{tabular}} & \multirow{2}{*}{\begin{tabular}[c]{@{}l@{}}Total\\ mAP\end{tabular}} \\
     & & & & & & & & & \\
    \midrule
    RTNH~\cite{Paek2022}         & R         & 49.9   & 56.7     & 52.8 & 42.0 & 41.5  & 50.6       & 44.5       & 47.4  \\
    \midrule
    Voxel-RCNN~\cite{Deng2021}   & L         & 81.8   & 69.6     & 48.8 & 47.1 & 46.9  & 54.8       & 37.2       & 46.4  \\
    CasA~\cite{Wu2022}           & L         & 82.2   & 65.6     & 44.4 & 53.7 & 44.9  & 62.7       & 36.9       & 50.9  \\
    TED-S~\cite{Wu2023}          & L         & 74.3   & 68.8     & 45.7 & 53.6 & 44.8  & 63.4       & 36.7       & 51.0  \\
    \midrule
    VPFNet~\cite{Zhu2023}        & C + L     & 81.2   & 76.3     & 46.3 & 53.7 & 44.9  & 63.1       & 36.9       & 52.2  \\
    TED-M~\cite{Wu2023}          & C + L     & 77.2   & 69.7     & 47.4 & 54.3 & 45.2  & 64.3       & 36.8       & 52.3  \\
    MixedFusion~\cite{Zhang2024} & C + L     & \textbf{84.5}   & \textbf{76.6}     & 53.3 & \textbf{55.3} & 49.6  & \textbf{68.7}       & 44.9       & 55.1  \\
    \midrule
    EchoFusion~\cite{LiuY2023}   & C + R     & 51.5      & 65.4        & 55.0    & 43.2    & 14.2     & 53.4          & 40.2          & 47.4  \\
    DPFT (ours)                  & C + R     & 55.7   & 59.4     & \textbf{63.1} & 49.0 & \textbf{51.6}  & 50.5 & \textbf{50.5} & \textbf{56.1}  \\
    \bottomrule
  \end{tabular}
\end{table*}

Since the published version of EchoFusion~\cite{LiuY2023} was only evaluated on the first 20 scenes of the K-Radar dataset, limited to a field of view (FoV) of \textpm\SI{20}{\degree} (instead of \textpm\SI{50}{\degree}) and did not use the official evaluation script, we retrained the EchoFusion~\cite{LiuY2023} model on the full dataset and evaluated it in accordance with the official evaluation scheme. All other results are in line with the literature.

The results of Table~\ref{tab:benchmark} show that our Dual Perspective Fusion Transformer achieves state-of-the-art performance on the challenging K-Radar dataset. The DPFT model achieves a mean average precision (mAP) value of \SI{56.1}{\percent} at an intersection over union (IoU) threshold of \SI{0.3}{} for 3D bounding box detection across all scene types. To account for any non-deterministic training behavior, the model is trained multiple times with different random seeds, such that \SI{56.1}{\percent} represents the mean across three runs with a standard deviation of \SI{1.1}{\percent}. Our proposed camera-radar fusion model outperforms both the radar-only RTNH~\cite{Paek2022} baseline model as well as the recently proposed EchoFusion~\cite{LiuY2023} camera-radar fusion. In comparison to state-of-the-art lidar or camera-lidar fusion models, it shows a significantly lower performance in normal conditions but outperforms them in particularly difficult weather conditions like fog, sleet, or heavy snow. This is most likely due to the radar's lower spatial resolution but higher robustness against environmental influences.

% \begin{figure}[t]
%   \centering
%   \includegraphics[width=\columnwidth]{figures/20240710_Fig_DPFT_Input_Modalities_v02.pdf}
%   \caption{Comparison of our DPFT model for different input modalities on the test data of the K-Radar dataset revision v2.0. The subscripts AE and RA stand for azimuth-elevation and range-azimuth and describe the utilization of just a single input perspective, whereas R is the combination of R$_{\text{AE}}$ and R$_{\text{RA}}$.}
%   \label{fig:multimodality}
% \end{figure}

\begin{table}[t]
  \caption{3D object detection results for different input modalities on the test data of the K-Radar dataset revision v2.0. The subscripts AE and RA describe the usage of just a single input perspective, namely azimuth-elevation or range-azimuth.}
  \label{tab:multimodality}
  \centering
  \begin{tabular}{lccccccc}
    \toprule
        & C   & R$_{\text{AE}}$ & R$_{\text{RA}}$ & R    & C + R$_{\text{AE}}$ & C + R$_{\text{RA}}$  & C + R \\
    \midrule
    mAP & 8.9 & 4.4             & 35.0            & 36.2 & 11.1                & 48.5                 & 50.5  \\
    \bottomrule
  \end{tabular}
\end{table}

The comparison of different sensor modalities, as shown in Table~\ref{tab:multimodality}, provides evidence for the effectiveness of our sensor fusion approach. It can be shown that the detection quality of the sensor fusion method exceeds even the combined performance of the individual sensor modalities, thus highlighting the effective use of the complementary sensor features. While the camera-only (C) configuration is similar to DETR3D~\cite{Wang2022} it struggles with the multitude of severe weather scenarios, the small backbone size, and the inability to utilize multi-view camera images. The results for the fusion of camera data with radar data from the range-azimuth (RA) plane in comparison to the fusion with data from the azimuth-elevation (AE) plane demonstrate the importance of the different perception planes for 3D object detection. However, the results with both radar perspectives, in comparison to only one perspective, suggest that the correspondence of the radar data and the camera data in the image plane, in combination with the physical relationship between the two radar perspectives, supports the fusion of the two sensor modalities. This shows the importance of the complementary information from the RA perception plane on one side and the benefits of the additional AE plane for the association between camera and radar on the other.

\subsection{Robustness}
\begin{figure}[t]
    \centering
    \vspace{-12pt}
    \subfloat[\empty]{%
        \includegraphics[width=0.485\linewidth]{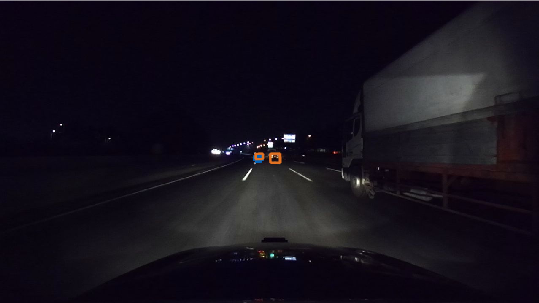}
    }
    \hfill
    \subfloat[\empty]{%
        \includegraphics[width=0.485\linewidth]{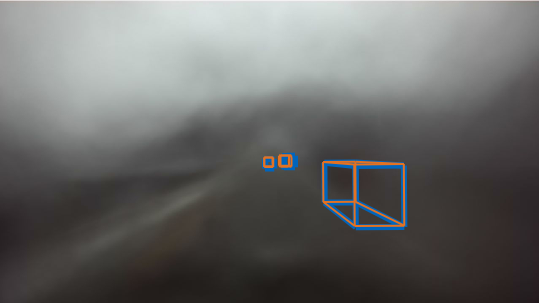}
    }
    \\ \vspace{-4pt}
    \subfloat[\empty]{%
        \includegraphics[width=0.485\linewidth]{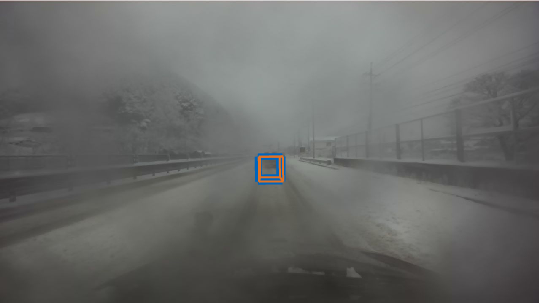}
    }
    \hfill
    \subfloat[\empty]{%
        \includegraphics[width=0.485\linewidth]{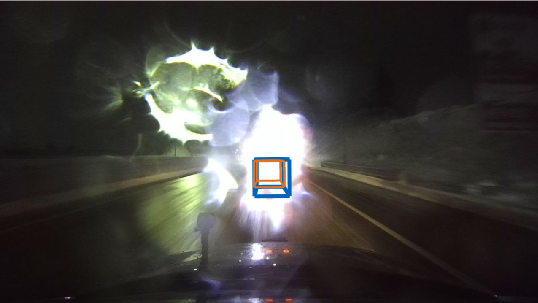}
    }
    \caption{Exemplary results of the model performance under night, rain, snow, and backlight conditions. The ground truth is shown in \textcolor{tumblue}{blue} and the model prediction in \textcolor{tumorange}{orange}.}
    \label{fig:robustness}
\end{figure}

The experimental results show the robustness of the DPFT model in two aspects: robustness against severe weather conditions and robustness against sensor failure. The robustness against severe weather conditions can be seen in Figure~\ref{fig:robustness} and shown by comparing the model performance under normal (norm.) conditions with the performance under different weather conditions of the K-Radar~\cite{Paek2022} dataset. As shown in Table~\ref{tab:benchmark}, the highest performance decrease for the DPFT model can be observed for the sleet condition, where a decrease of \SI{6.8}{\percent} can be measured in comparison to the normal condition. In comparison to that, the performance of the MixedFusion~\cite{Zhang2024} model decreased by \SI{41.3}{\percent} and the performance of EchoFusion~\cite{LiuY2023} decreased by \SI{76.3}{\percent}. In general, our proposed DPFT method shows an average performance difference of \SI{-2.5}{\percent} between the normal and all other conditions. In comparison, the RTNH~\cite{Paek2022}, MixedFusion~\cite{Zhang2024}, and EchoFusion~\cite{LiuY2023} models show a decrease of \SI{-3.8}{\percent}, \SI{-31.3}{\percent}, and \SI{-12.8}{\percent}, respectively.
The analysis of the average and maximum decrease suggests that models that are considering radar data are less affected by varying weather conditions than those that are not considering radar data. Ultimately, it can be shown that our proposed DPFT model shows high robustness against server weather conditions and is equally robust as the radar-only RTNH~\cite{Paek2022} method. However, the unimodal RTNH~\cite{Paek2022} model performance is significantly lower and it cannot deal with a sensor modality failure.

\begin{table}[t]
    \caption{Results with simulated sensor failure on the test set of the K-Radar dataset revision v2.0.}
    \label{tab:failure}
    \centering
    \begin{tabular}{@{}p{0.13\linewidth}p{0.13\linewidth}p{0.16\linewidth}p{0.18\linewidth}p{0.18\linewidth}@{}}
        \toprule
        Trained \hfil & Tested \hfil & mAP \hfil & mAP$_{\text{pre-trained}}$ \hfil & mAP$_{\text{dropout}}$ \hfil  \\
        \midrule
        C \hfil    & C \hfil & 8.9 \hfil  & - \hfil  & - \hfil   \\
        R \hfil    & R \hfil & 36.2 \hfil & - \hfil  & - \hfil   \\
        C + R \hfil & C \hfil & 1.1 \hfil & 0.0 \hfil & 9.2 \hfil   \\
        C + R \hfil & R \hfil & 11.1 \hfil & 12.8 \hfil & 37.5 \hfil   \\
        C + R \hfil & C + R \hfil & 50.5 \hfil & 51.4 \hfil & 38.3 \hfil   \\
        \bottomrule
    \end{tabular}
\end{table}

\begin{table*}[t]
    \caption{Performance and complexity for different backbones and input image resolutions.}
    \label{tab:complexity}
    \centering
    \begin{tabular}{@{}p{0.18\linewidth}p{0.09\linewidth}p{0.09\linewidth}p{0.09\linewidth}p{0.09\linewidth}p{0.09\linewidth}p{0.09\linewidth}@{}}
         \toprule
         & \hfil ResNet101 & \hfil ResNet50 & \hfil ResNet34 & \hfil 720px & \hfil 512px & \hfil 256px  \\
        \midrule
        mAP in \% &  \hfil 50.5 & \hfil 49.8 & \hfil 47.2 & \hfil 50.5 & \hfil 50.5 &  \hfil 45.4 \\
        Inference time in ms &  \hfil 87 & \hfil 69 & \hfil 64 & \hfil 94 & \hfil 87 & \hfil 81 \\
        Complexity in TFLOPs &  \hfil 0.16 & \hfil 0.09 & \hfil 0.08 & \hfil 0.30 & \hfil 0.16 & \hfil 0.04 \\
        \bottomrule
    \end{tabular}
\end{table*}

The robustness of our method against sensor failure is achieved by a model design without interdependencies between the different modalities. While this prevents a complete failure of the model if a single sensor modality fails during runtime, the model performance still drops significantly, as shown in Table~\ref{tab:failure}. To counteract this, we used the pre-trained weights of the camera and radar-only models as initialization~\cite{Liang2022}, but could not observe any significant changes. Besides that, we trained the model with modality dropout~\cite{Ge2023} and were able to improve the performance for the sensor failure cases, but observed a significant decrease under nominal conditions, which is in contrast to~\cite{Ge2023, Wang2023}.

\subsection{Complexity}
Our model is designed for real-world applications, which is why inference time and memory consumption measurements are conducted. All tests are executed on a dedicated benchmark sever equipped with an NVIDIA V100 GPU and isolated in a containerized environment. The DeepSpeed~\cite{Rasley2020} framework is used for reliable and accurate measurements.

The proposed DPFT model achieves an inference time of \SI{87}~\textpm\SI{1}{\milli\second}, which is lower than the \SI{100}{\milli\second} cycle time of the radar sensor. This is important to be able to process every sensor image and not have to drop any. In comparison, MixedFusion~\cite{Zhang2024}, and EchoFusion~\cite{LiuY2023} have an inference time of \SI{143}{\milli\second}, and \SI{348}{\milli\second}, respectively. Therefore, our DPFT model achieves the lowest inference time among all tested~methods.

The overall model complexity is mainly driven by the backbone selection, while the memory consumption is mainly caused by the input image size, as shown in Table~\ref{tab:complexity}. The baseline implementation of or DPFT model requires \SI{4.0}{\gibi\byte} of GPU memory during inference and has a measured computational complexity of \SI{0.16}{\tera\flops}. In comparison, EchoFusion~\cite{LiuY2023} requires \SI{3.5}{\gibi\byte} of GPU memory, but has a computational complexity of \SI{1.52}{\tera\flops}, explaining its higher inference time. This comparison shows the computational efficiency of our proposed method even without any runtime optimization like TensorRT. Moreover, the modular design of our implementation allows the usage of different backbones and input image sizes. As shown in Table~\ref{tab:complexity}, altering these parameters can significantly decrease the computational complexity but influence the model performance. As a consequence, these parameters have to be chosen in accordance with the desired application.

\subsection{Ablation Study}
\begin{figure}[t]
  \centering
  \includegraphics[width=\columnwidth,trim={0 0.04cm 0 0.02cm},clip]{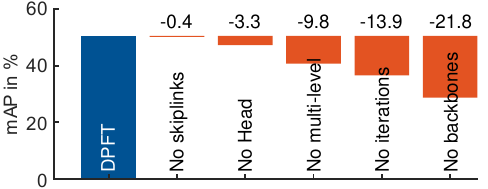}
  \caption{Performance loss due to the ablation of individual model components on the test data of the K-Radar dataset revision v2.0.}
  \label{fig:ablation}
\end{figure}

The results of the ablation study show the contribution of the individual model components on the overall detection performance and are shown in Figure~\ref{fig:ablation}. It can be seen that the ablation of the backbones causes the greatest performance decreases, whereas the contribution of the skiplinks is not significant. Besides that, the iterative refinement process and the usage of multi-level feature maps have a significant effect on the detection performance. Moreover, in consideration of the conducted experiments on the input data modalities (Table~\ref{tab:multimodality}) and the analysis of different backbones (Table~\ref{tab:complexity}), the results suggest that the sensor fusion is the most important factor for the model performance.

\subsection{Discussion}
\begin{figure}[t]
    \centering
    \subfloat[\empty]{%
        \includegraphics[clip, trim=0.0cm 1.2cm 0.0cm 0.8cm, width=0.985\linewidth]{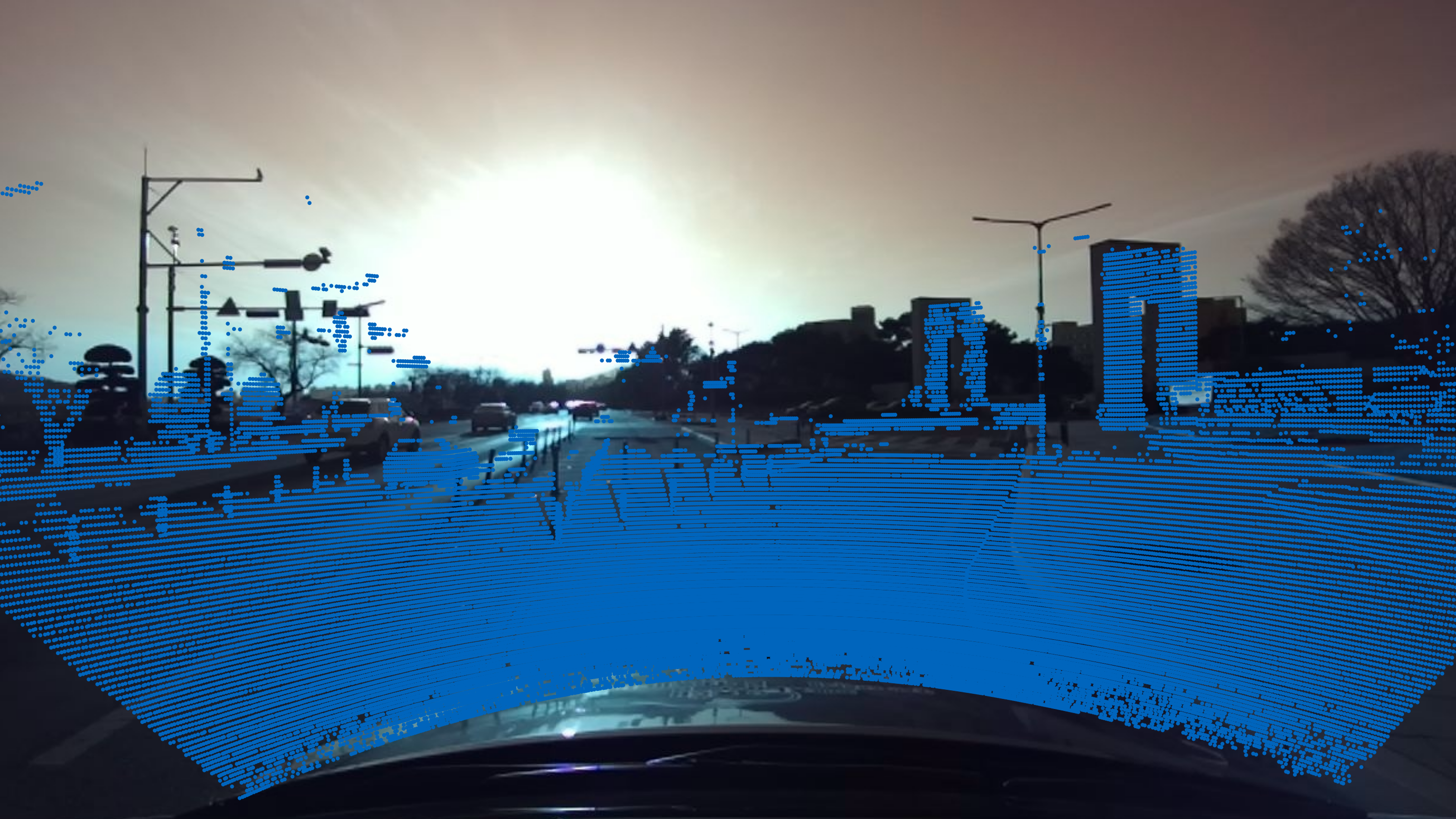}
    }
    \\ \vspace{-4pt}
    \subfloat[\empty]{%
        \includegraphics[width=0.485\linewidth]{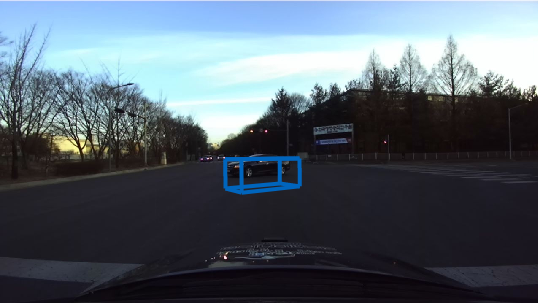}
    }
    \hfill
    \subfloat[\empty]{%
        \includegraphics[width=0.485\linewidth]{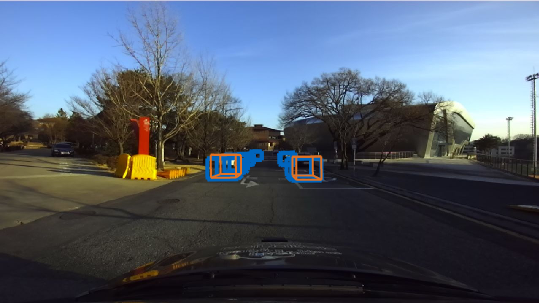}
    }
    \caption{Visualization of the dataset's sensor miscalibration (left) and two failure cases of the model. One shows a missing detection of a crossing object (center) and the other shows false negatives for partially occluded objects (right). The ground truth is shown in \textcolor{tumblue}{blue} and the model prediction in \textcolor{tumorange}{orange}.}
    \label{fig:discussion}
\end{figure}

While our model achieved state-of-the-art results in the conducted experiments, there are certain limitations to it. Firstly, it outperforms lidar and camera-lidar fusion methods only in severe weather conditions while showing a significantly lower performance in normal conditions. Secondly, the model has difficulties detecting objects that are moving tangential to the ego vehicle's direction of travel and correctly predicting their heading angle, as shown in Figure~\ref{fig:discussion}. This is probably caused by the fact that crossing objects are heavily underrepresented in the dataset on the one side and the inability of the radar sensor to measure tangential velocities on the other. Furthermore, it struggles to detect or differentiate between multiple objects that are behind each other or close to each other, which can be seen in Figure~\ref{fig:discussion}. We believe that this is due to the partial occlusion and the limited resolution of the radar sensor in the azimuth-elevation plane. Last but not least, the generalization capability of the model could only be tested within the scope of the K-Radar~\cite{Paek2022} dataset. Since the K-Radar~\cite{Paek2022} dataset is the only dataset that provides raw 4D radar data for different weather conditions and the only large-scale dataset with radar cube data in general, the transferability of the model to different datasets is yet to be shown. Nevertheless, comparable model architectures~\cite{LiuY2023} that only rely on 3D radar data show promising generalization results, which is a first indicator for the transferability of these model types.

Despite being the only dataset with 4D radar cube data, the K-Radar~\cite{Paek2022} dataset shows some labeling inconsistencies (especially between the sedan and bus or truck classes) even within the revision v2.0. In addition, the test set is sampled from the same driving sequences and contains similar scenarios to the train set, which limits the ability to test the generalizability of models, even if the test split is formally independent. Furthermore, we observed a misalignment between the camera and lidar frame, as shown in Figure~\ref{fig:discussion}, which is important because the labels are created on the lidar data, and which is why EchoFusion~\cite{LiuY2023} used their own calibration. However, a recalibration of the sensors is difficult and would limit the comparability to previous methods, which is why we used the official calibration. Nevertheless, further investigations would be needed to quantify the model's sensitivity to miscalibrations. Last but not least, the calculation of the total mAP metric in the official evaluation scheme could be misleading since it is calculated as the weighted average of the individual categories weighted by the number of ground truth objects. In general, the usage of the KITTI~\cite{Geiger2012} evaluation protocol could be questioned due to the problem of average precision distortion \cite{Zhang2022} and since recent studies show that other metrics, like the nuScenes detection score (NDS), correlate better with the fulfillment of the autonomous driving task~\cite{Schreier2023}.

\section{Conclusion}
\label{sec:conclusion}
We proposed a novel method to fuse camera and cube-level radar data to achieve a performant, robust, yet cost-effective method for 3D object detection. We are the first to fuse raw 4D radar data with camera data and demonstrate the importance of the different input perspectives. Our proposed DPFT method achieves state-of-the-art results in the challenging environmental conditions of the K-Radar dataset. Experimental results show that our proposed method is robust against severe weather conditions and is able to maintain general functionality even after a sensor failure. Finally, we provided a comprehensive analysis of the computational complexity of our method and were able to show that our method has the fastest inference time among all tested fusion methods.

Despite the great potential of camera and radar fusion for 3D object detection, new research questions emerge from this work. While we proposed a novel dual perspective fusion approach, the general question of how to utilize the high dimensional radar data most efficiently remains open for research. Moreover, balancing the performance of different sensor modalities within a fusion method to exploit the input data most effectively and avoid significant performance losses during the event of a sensor failure remains challenging. Even if we used different methods to counteract the performance degradation after a sensor failure, further research is needed to mitigate this effect. Moreover, sensor-specific challenges like target separation in the radar domain or depth estimation in the camera domain remain open for research. Beyond that, temporal information could be considered to increase the performance and a different detection head could be used to realize an instance-free detection method in future work.

% References Section
\bibliographystyle{IEEEtran}
\bibliography{references}

% \newpage

\section{Biography Section}
 
\vspace{11pt}

\vspace{-33pt}
\begin{IEEEbiography}[{\includegraphics[width=1in,clip,keepaspectratio]{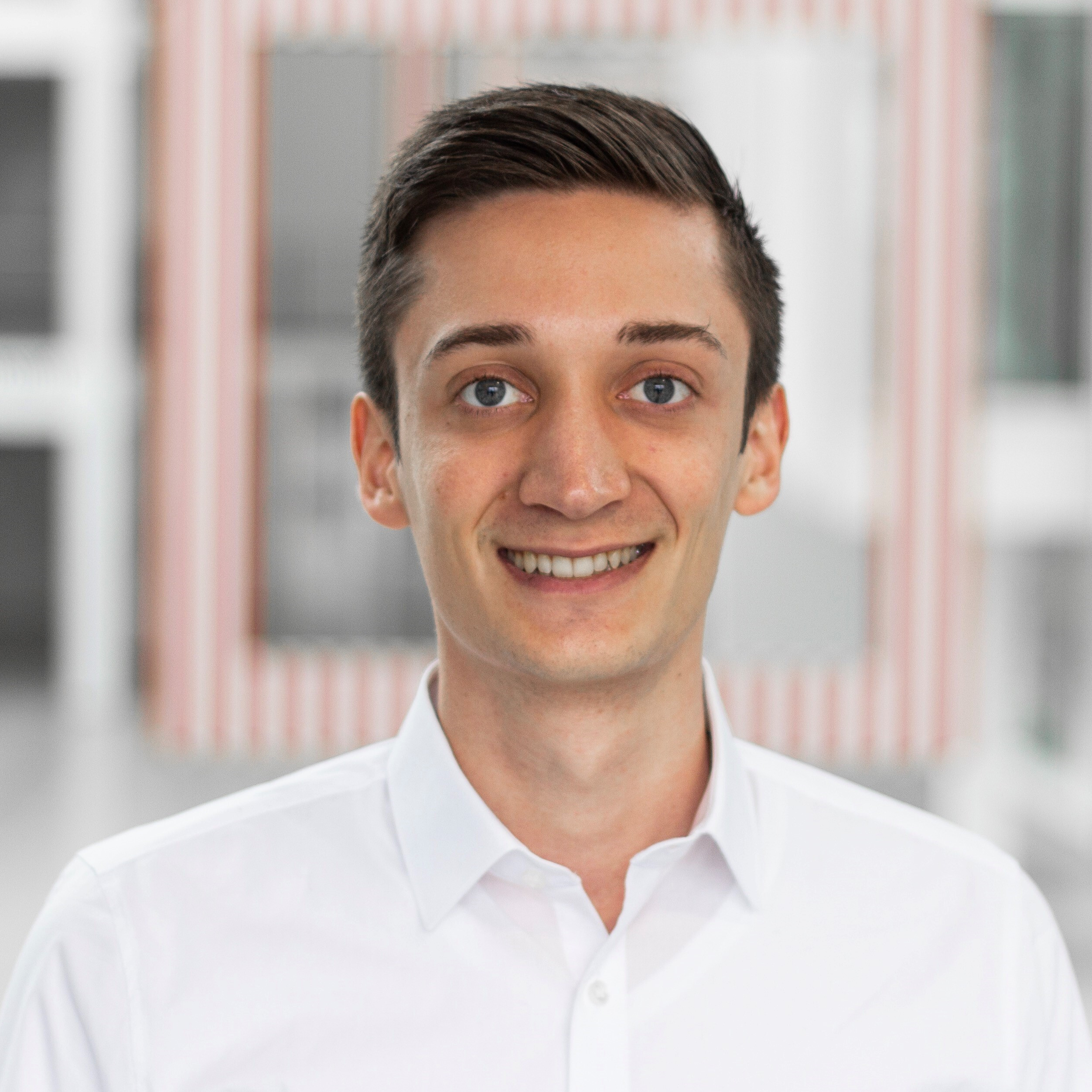}}]{Felix Fent}
    received the B.Sc. and M.Sc. degrees from the Technical University of Munich (TUM), Munich, Germany, in 2018 and 2020, respectively, where he is currently pursuing a Ph.D. degree in
    mechanical engineering with the Institute of Automotive Technology. His research interests include radar-based perception, sensor fusion and multi-modal object detection approaches with a focus on real-world applications.
\end{IEEEbiography}

\vspace{11pt}

\vspace{-33pt}

\begin{IEEEbiography}[{\includegraphics[width=1in,clip,keepaspectratio]{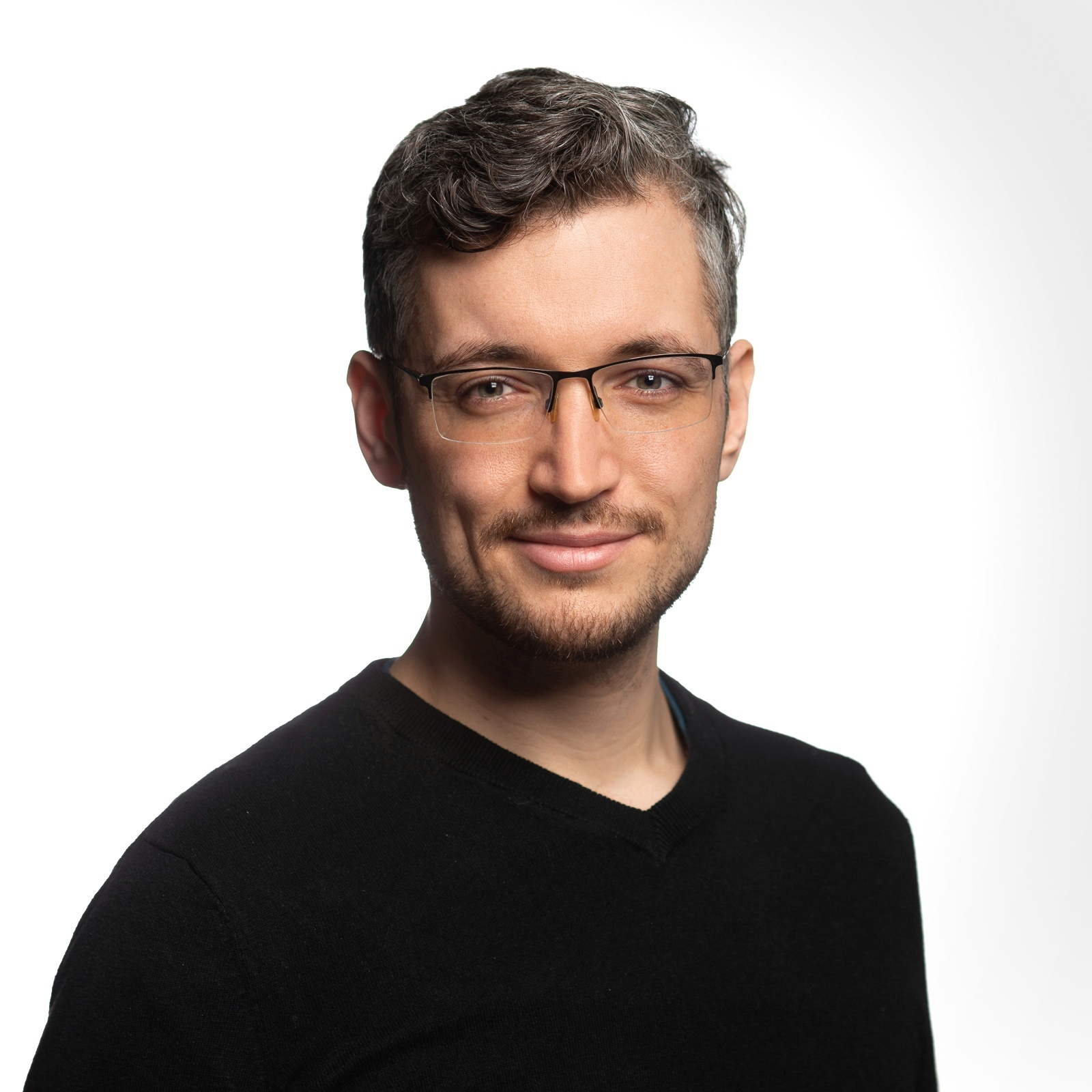}}]{Andras Palffy}
(Member, IEEE) received the M.Sc. degree in computer science engineering from Pazmany Peter Catholic University, Budapest, Hungary, in 2016, and the M.Sc. degree in digital signal and image processing from Cranfield University, Cranfield, U.K., in 2015. From 2013 to 2017, he was an algorithm researcher at Eutecus, a US based startup developing computer vision algorithms for traffic monitoring and driver assistance applications. He obtained his Ph.D. degree in 2022 at Delft University of Technology, Delft, Netherlands, focusing on radar based vulnerable road user detection for automated driving. In 2022 he co-founded Perciv AI, a machine perception startup developing AI-driven, next generation machine perception for radars.
\end{IEEEbiography}

\vspace{11pt}

\vspace{-33pt}

\begin{IEEEbiography}
[{\includegraphics[width=1in,clip,keepaspectratio]{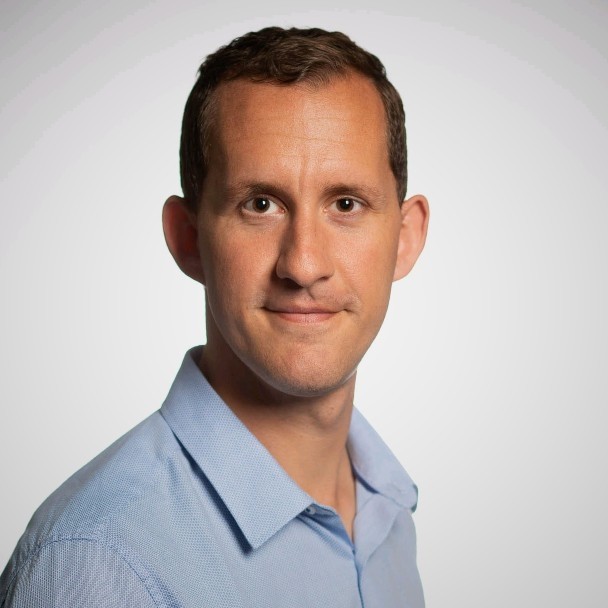}}]{Holger Caesar}
is an Assistant Professor in the Intelligent Vehicles group of TU Delft in the Netherlands. Holger's research interests are in the area of Autonomous Vehicle perception and prediction, with a particular focus on scalability of learning and annotation approaches. Previously he was a Principal Research Scientist at Motional. There he started three teams with 20+ members that focused on Data Annotation, Autolabeling and Data Mining. Holger received a PhD in Computer Vision from the University of Edinburgh, Scotland and studied in Germany and Switzerland (KIT Karlsruhe, EPF Lausanne, ETH Zurich). He is best known for developing the influential autonomous driving datasets nuScenes and nuPlan, as well as his contributions to the real-time 3d object detection method PointPillars.
\end{IEEEbiography}

\vfill

\newpage

{\appendices
\include{appendix}}

\end{document}

%% file: appendix.tex
\addcontentsline{toc}{section}{Appendices}
\newcommand{\hbAppendixPrefix}{A}
\renewcommand{\thefigure}{\hbAppendixPrefix\arabic{figure}}
\setcounter{figure}{0}
\renewcommand{\thetable}{\hbAppendixPrefix\arabic{table}} 
\setcounter{table}{0}

\section{Additional Result Details}
The appendix presents additional details on the results on the K-Radar~\cite{Paek2022} dataset. Following Section~\ref{sec:results}, the results of Table~\ref{tab:appx_iou} were obtained on the original version (revision v1.0) of the dataset, while all other additional results are based on the revised version (revision v2.0) of the dataset.

\begin{table*}[t]
  \caption{Object detection results for the K-Radar test set revision v1.0.}
  \label{tab:appx_iou}
  \centering
  \begin{tabular}{@{}p{0.12\textwidth}p{0.06\textwidth}p{0.08\textwidth}p{0.08\textwidth}p{0.08\textwidth}p{0.005\textwidth}p{0.08\textwidth}p{0.08\textwidth}p{0.08\textwidth}@{}}
    \toprule
     &  & \multicolumn{3}{c}{3D mAP} & & \multicolumn{3}{c}{BEV mAP} \\
    \cline{3-5} \cline{7-9}
    Method & Modality & \hfil AP@0.3 & \hfil AP@0.5 & \hfil AP@0.7 & & \hfil AP@0.3 & \hfil AP@0.5 & \hfil AP@0.7 \\
    \midrule
    RTNH~\cite{Paek2022}         & R         & \hfil 47.4 & \hfil 15.6 & \hfil 0.5 & & \hfil \textbf{58.4} & \hfil 43.2 & \hfil 11.5  \\
    EchoFusion~\cite{LiuY2023}   & C + R     & \hfil 47.4 & \hfil 28.1 & \hfil 6.4 & & \hfil 48.9 & \hfil 39.7 & \hfil 25.7  \\
    DPFT (ours)                  & C + R     & \hfil \textbf{56.1} & \hfil \textbf{37.0} & \hfil \textbf{8.0} & & \hfil 57.5 & \hfil \textbf{48.5} & \hfil \textbf{26.3}  \\
    \bottomrule
  \end{tabular}
\end{table*}

The results of Table~\ref{tab:appx_iou} show decreasing mAP values with increasing IoU thresholds for all tested methods and both the 3D and the BEV object detection tasks. It is worth mentioning that the results of the DPFT method, listed in Table~\ref{tab:appx_iou}, are the mean values of three independent model trainings to mitigate the effects of any non-deterministic training behavior. It can be seen that our proposed method outperforms all previous radar- and camera-radar-based methods for 3D object detection at all IoU thresholds and performs on par with the RTNH~\cite{Paek2022} method for BEV detections at a low IoU threshold. However, our proposed method archives higher mAP values for BEV detection at higher IoU thresholds.

\begin{table}[h]
    \caption{3D object detection results for different detection ranges.}
    \label{tab:appx_distance}
    \centering
    \begin{tabular}{@{}p{0.12\linewidth}p{0.09\linewidth}p{0.12\linewidth}p{0.14\linewidth}p{0.14\linewidth}p{0.14\linewidth}@{}}
        \toprule
        Modality & \hfil Total & \hfil 0~-~\SI{10}{\meter} & \hfil 10~-~\SI{30}{\meter} & \hfil 30~-~\SI{50}{\meter} & \hfil 50~-~\SI{72}{\meter} \\
        \midrule
        C     & \hfil 8.9  & \hfil 27.3 & \hfil 15.5 & \hfil 4.7 & \hfil 3.4  \\
        R     & \hfil 36.2 & \hfil 35.5 & \hfil 42.7 & \hfil 37.1 & \hfil 25.2  \\
        C + R & \hfil 50.5 & \hfil 44.8 & \hfil 54.6 & \hfil 53.4 & \hfil 35.3  \\
        \bottomrule
    \end{tabular}
\end{table}

The experimental results of Table~\ref{tab:appx_distance} show the performance of our DPFT model for different sensor modalities and detection range bins. It can be seen that the general performance of the model decreases with increasing range. This is especially true for the camera-only model, which shows a significant performance decrease with increasing detection range. The observed behavior is probably caused by the inability of the camera sensor to measure depth information and its decreasing spatial resolution with increasing distance. In contrast, the radar-only model shows a lower performance for the range between 0~-~\SI{10}{\meter} and achieves the highest performance in a range between 10~-~\SI{30}{\meter}, with a decreasing performance over increasing distance. This phenomenon is probably caused by the higher noise level of the radar in close range and the decreasing spatial resolution with increasing distance. The performance of the camera-radar fusion model shows a similar behavior to the radar-based model, but a higher performance overall and seems to be less affected by increasing distance. We believe that this is a result of the already discussed sensor properties and the distribution of objects in the dataset that contains the most objects in a range of 20~-~\SI{40}{\meter} and the least for distances greater than \SI{60}{\meter}~\cite{Paek2022}.

\section{Additional Details on Robustness}
In addition to the differentiation into different weather conditions, the K-Radar dataset allows the separate determination of the performance values for day and night conditions. The results of Table~\ref{tab:appx_night} show that all configurations of the DPFT model perform better under daytime conditions than nighttime conditions. Nevertheless, the performance of the camera-only model is affected the most, while the radar-only model shows the smallest decrease of all tested configurations. This is probably because camera sensors are dependent on ambient light, while radar sensors are active sensors and, therefore, independent from external sources. However, the general tendency could also be explained by the data distribution of the K-Radar dataset, which consists of \SI{63}{\percent} daytime scenes, which results in an imbalanced training and test set~\cite{Paek2022}.

\begin{table}[h]
    \caption{3D object detection results for different daytimes.}
    \label{tab:appx_night}
    \centering
    \begin{tabular}{@{}p{0.12\linewidth}p{0.12\linewidth}p{0.12\linewidth}p{0.12\linewidth}p{0.12\linewidth}p{0.12\linewidth}@{}}
        \toprule
        Modality & \hfil Day & \hfil Night & \hfil Total \\
        \midrule
        C     & \hfil 9.8 & \hfil 3.0 & \hfil 8.9   \\
        R     & \hfil 36.9 & \hfil 29.1 & \hfil 36.2  \\
        C + R & \hfil 52.7 & \hfil 39.8 & \hfil 50.5  \\
        \bottomrule
    \end{tabular}
\end{table}

The analysis of individual models shows that the camera-based model fails if the camera lens is covered by raindrops or sleet (as shown in \ref{fig:appx_examples}), which only gets worse in night-time conditions. However, these problems cloud be avoided by a different camera positioning or cleaning mechanism. The radar-based performance seems to be less affected by environmental conditions but more dependent on the number of available training samples. Nevertheless, target separation remains challenging in dense traffic or city scenarios.

\section{Additional Details on Complexitiy}
In this section, we provide more detailed results on the model complexity analysis discussed in Section~\ref{sec:results}. The appended Table~\ref{tab:appx_complexity} is an extension of Table~\ref{tab:complexity} and includes additional metrics on the computational complexity of the different model configurations as well as the memory requirements based on the model parameters. In general, it provides evidence for the claim that a larger backbone size and higher input resolution lead to a higher model performance but an increased computational complexity.

\begin{table*}[t]
    \caption{Performance and complexity for different backbones and input image resolutions.}
    \label{tab:appx_complexity}
    \centering
    \begin{tabular}{@{}p{0.16\linewidth}p{0.09\linewidth}p{0.09\linewidth}p{0.09\linewidth}p{0.09\linewidth}p{0.09\linewidth}p{0.09\linewidth}@{}}
         \toprule
         & \hfil ResNet101 & \hfil ResNet50 & \hfil ResNet34 & \hfil 720px & \hfil 512px & \hfil 256px  \\
        \midrule
        mAP in \% &  \hfil 50.5 & \hfil 49.8 & \hfil 47.2 & \hfil 50.5 & \hfil 50.5 &  \hfil 45.4 \\
        Time in ms &  \hfil 87 & \hfil 69 & \hfil 64 & \hfil 94 & \hfil 87 & \hfil 81 \\
        FLOPs in \SI{e9}{} &  \hfil 156 & \hfil 86 & \hfil 75 & \hfil 302 & \hfil 156 & \hfil 44 \\
        MACs in \SI{e9}{} &  \hfil 78 & \hfil 43 & \hfil 37 & \hfil 150 & \hfil 78 & \hfil 22 \\
        Parameters in \SI{e6}{} &  \hfil 90 & \hfil 66 & \hfil 44 & \hfil 90 & \hfil 90 & \hfil 90 \\
        \bottomrule
    \end{tabular}
\end{table*}

In addition, the model has been tested with different numbers of query points to analyze the effects of different query point resolutions on the model performance and computational complexity. The results for 100, 400, and 900 query points show that an increased query point resolution leads only to a marginal increase in computational complexity of \SI{156}{}, \SI{156}{}, and \SI{157}{\giga\flops}, but a larger impact on the memory consumption. In contrast, the best model performance seems to be achieved with 400 query points, whereas a query point resolution of 100 and 900 leads to a result of \SI{47.8}{\percent} and \SI{44.5}{\percent} mAP, respectively. During model development, quadratic and exponentially distributed query point initializations in both cartesian and polar coordinates as well as a learnable query point initialization were also tested with no significant performance increases. Besides that, Table~\ref{tab:appx_speed} provides inference time measurements on different hardware accelerators, using the same method as described in Section~\ref{sec:results} and demonstrates that significantly lower inference times can be achieved on more modern GPUs.

\begin{table}[h]
    \caption{Inference time on different NVIDIA GPU units.}
    \label{tab:appx_speed}
    \centering
    \begin{tabular}{@{}p{0.17\linewidth}p{0.11\linewidth}p{0.11\linewidth}p{0.11\linewidth}p{0.11\linewidth}p{0.11\linewidth}@{}}
         \toprule
         & \hfil 3090 & \hfil 4090 & \hfil V100 & \hfil A40  & \hfil A100 \\
        \midrule
        Time in ms &  \hfil 74\textpm1.2 & \hfil 32\textpm0.4 & \hfil 87\textpm1.2 & \hfil 52\textpm1.0 & \hfil 41\textpm0.1 \\
        \bottomrule
    \end{tabular}
\end{table}

\section{Examples}
\begin{figure*}[t]
    \centering
    \vspace{-12pt}
    \subfloat[\empty]{%
        \includegraphics[width=0.32\linewidth]{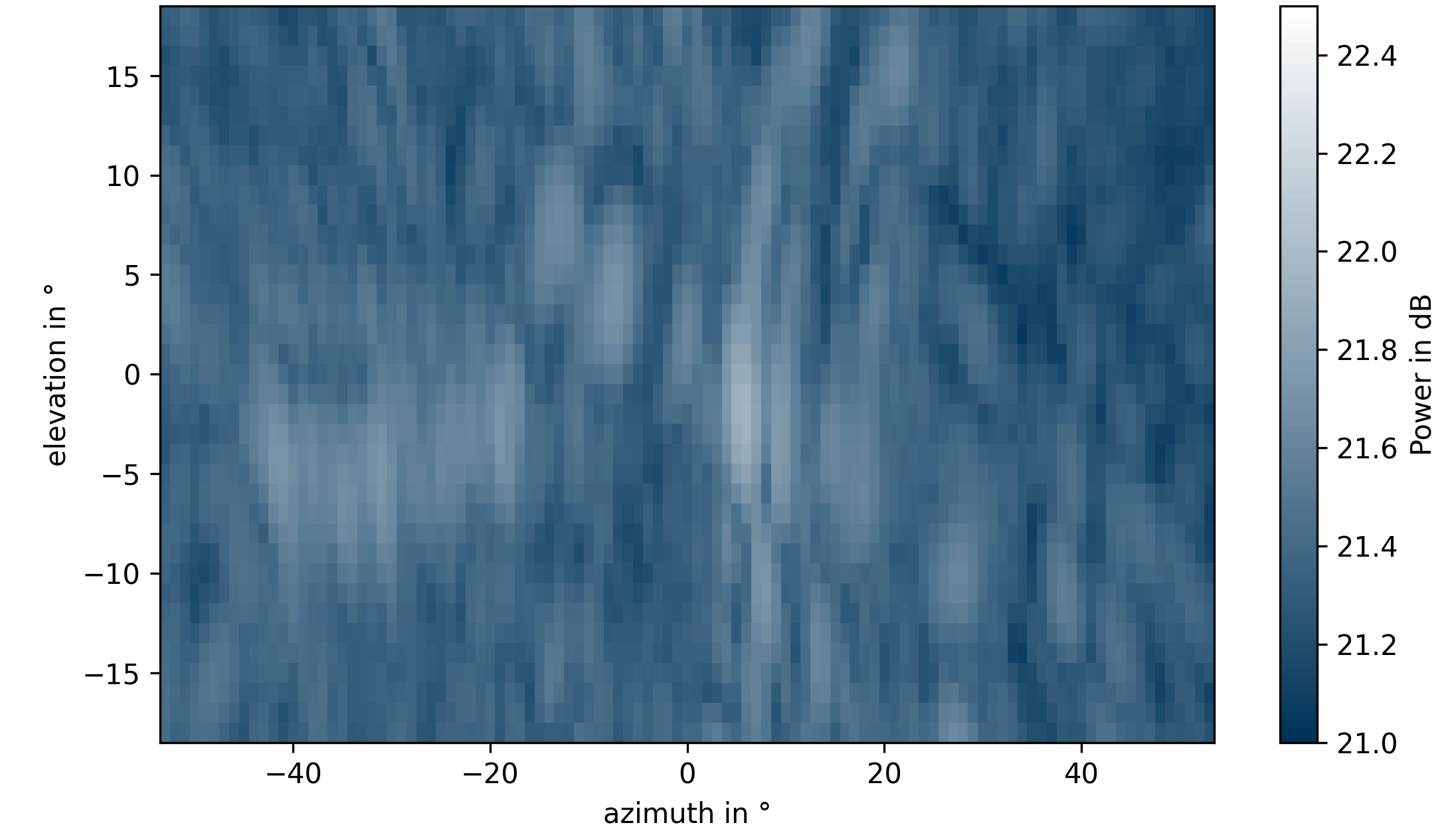}
    }
    \hfill
    \subfloat[\empty]{%
        \includegraphics[width=0.32\linewidth]{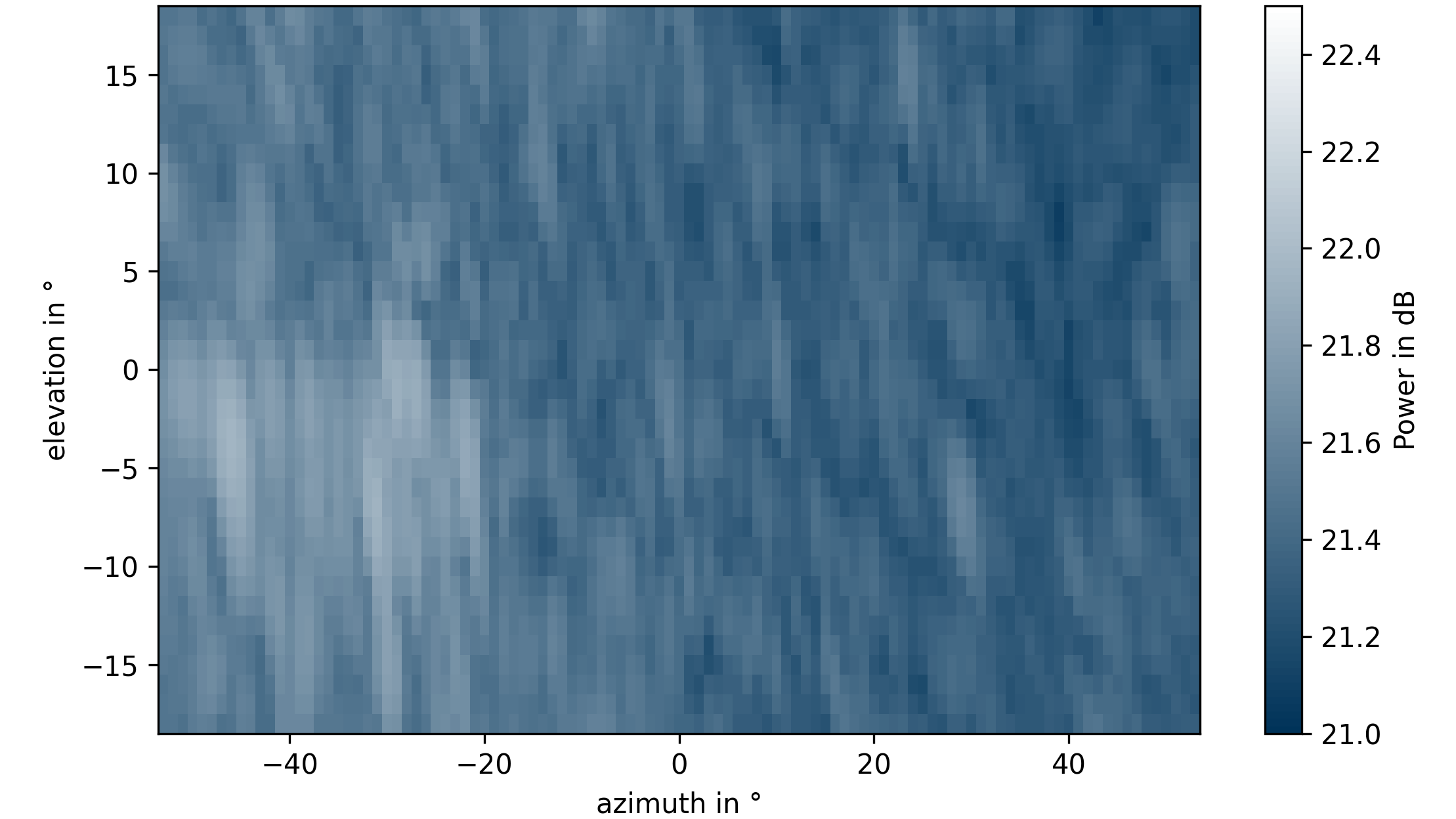}
    }
    \hfill
    \subfloat[\empty]{%
        \includegraphics[width=0.32\linewidth]{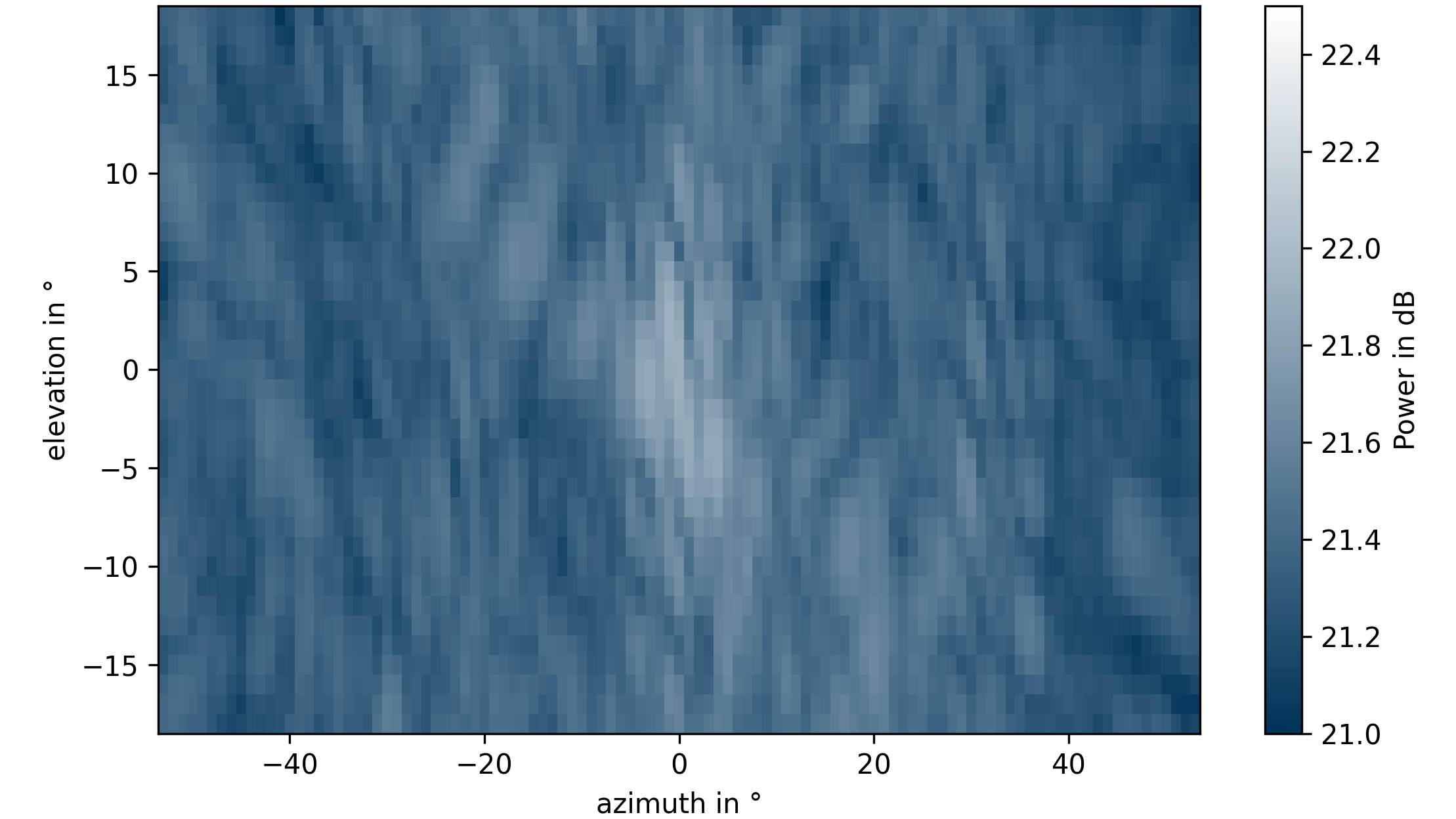}
    }
    \\ \vspace{-4pt}
    \subfloat[\empty]{%
        \includegraphics[width=0.25\linewidth]{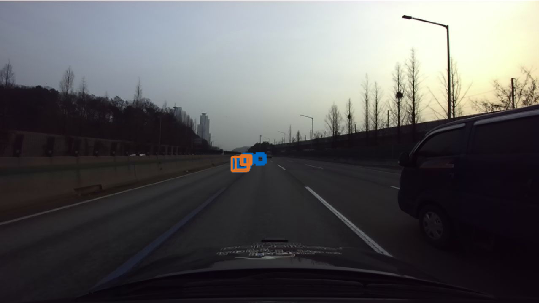}
    }
    \hspace{38pt}
    \subfloat[\empty]{%
        \includegraphics[width=0.25\linewidth]{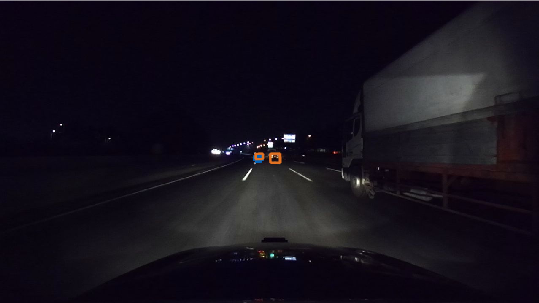}
    }
    \hspace{38pt}
    \subfloat[\empty]{%
        \includegraphics[width=0.25\linewidth]{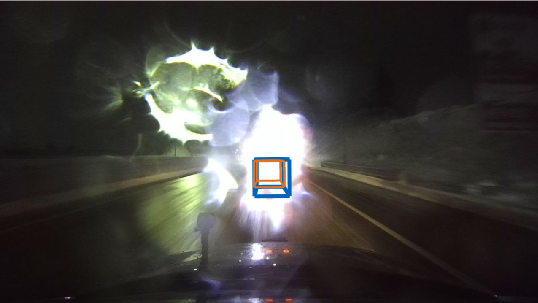}
    }
    \setcounter{subfigure}{0}
    \\ \vspace{-4pt}
    \subfloat[dusk]{%
        \includegraphics[width=0.32\linewidth]{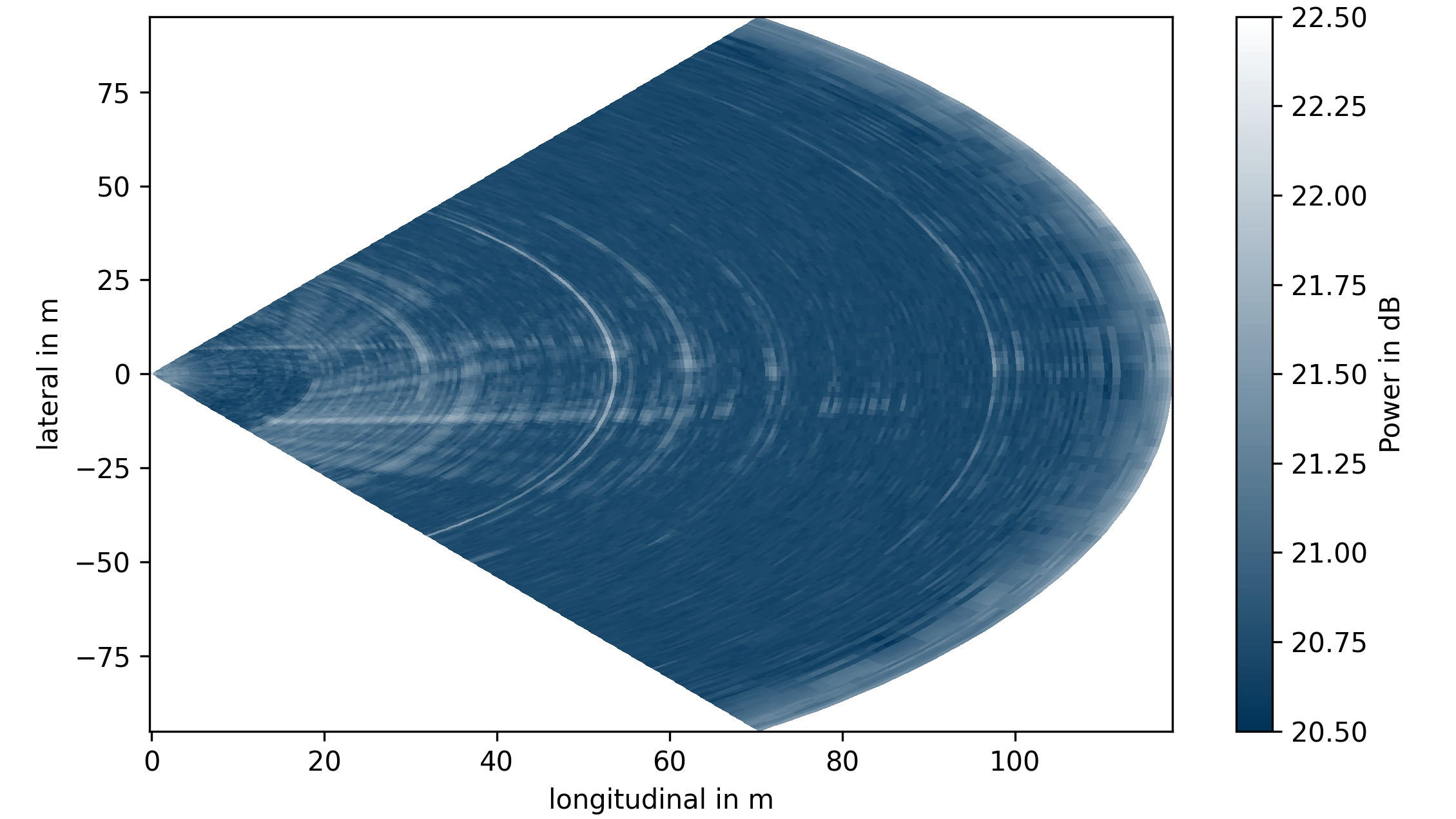}
    }
    \hfill
    \subfloat[\empty][night]{%
        \includegraphics[width=0.32\linewidth]{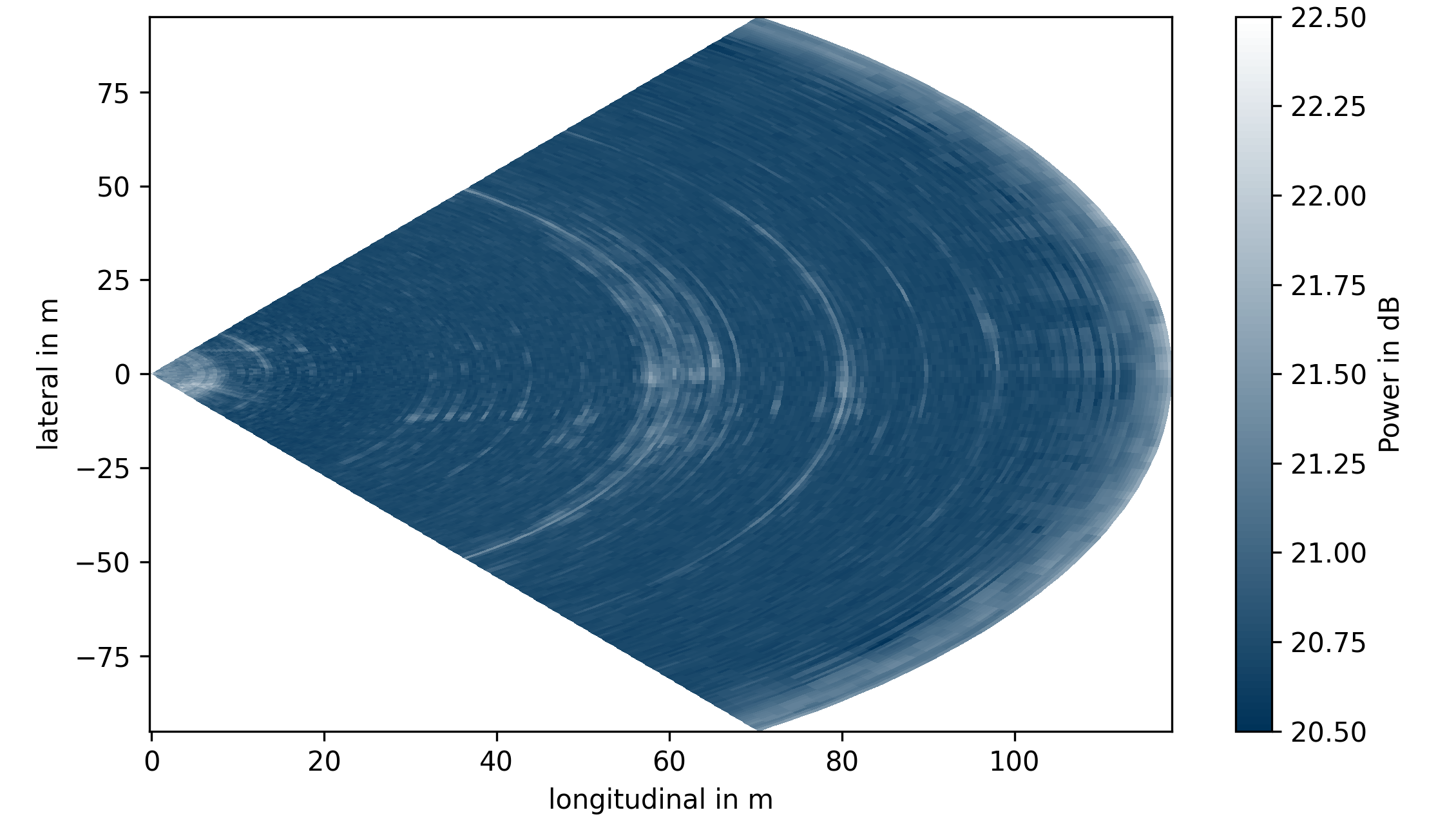}
    }
    \hfill
    \subfloat[\empty][blinded]{%
        \includegraphics[width=0.32\linewidth]{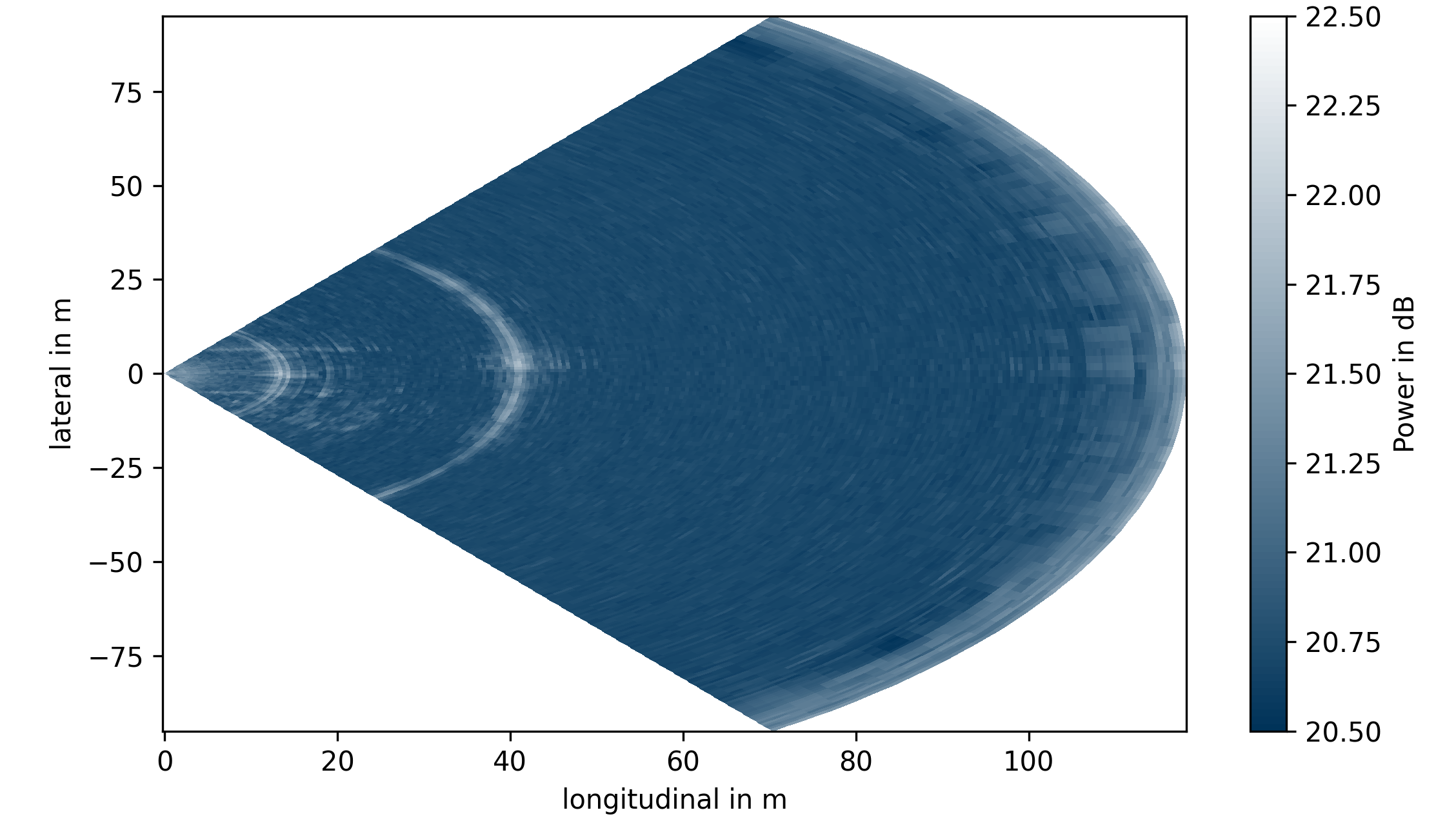}
    }
    \\ \vspace{-4pt}
    \subfloat[\empty]{%
        \includegraphics[width=0.32\linewidth]{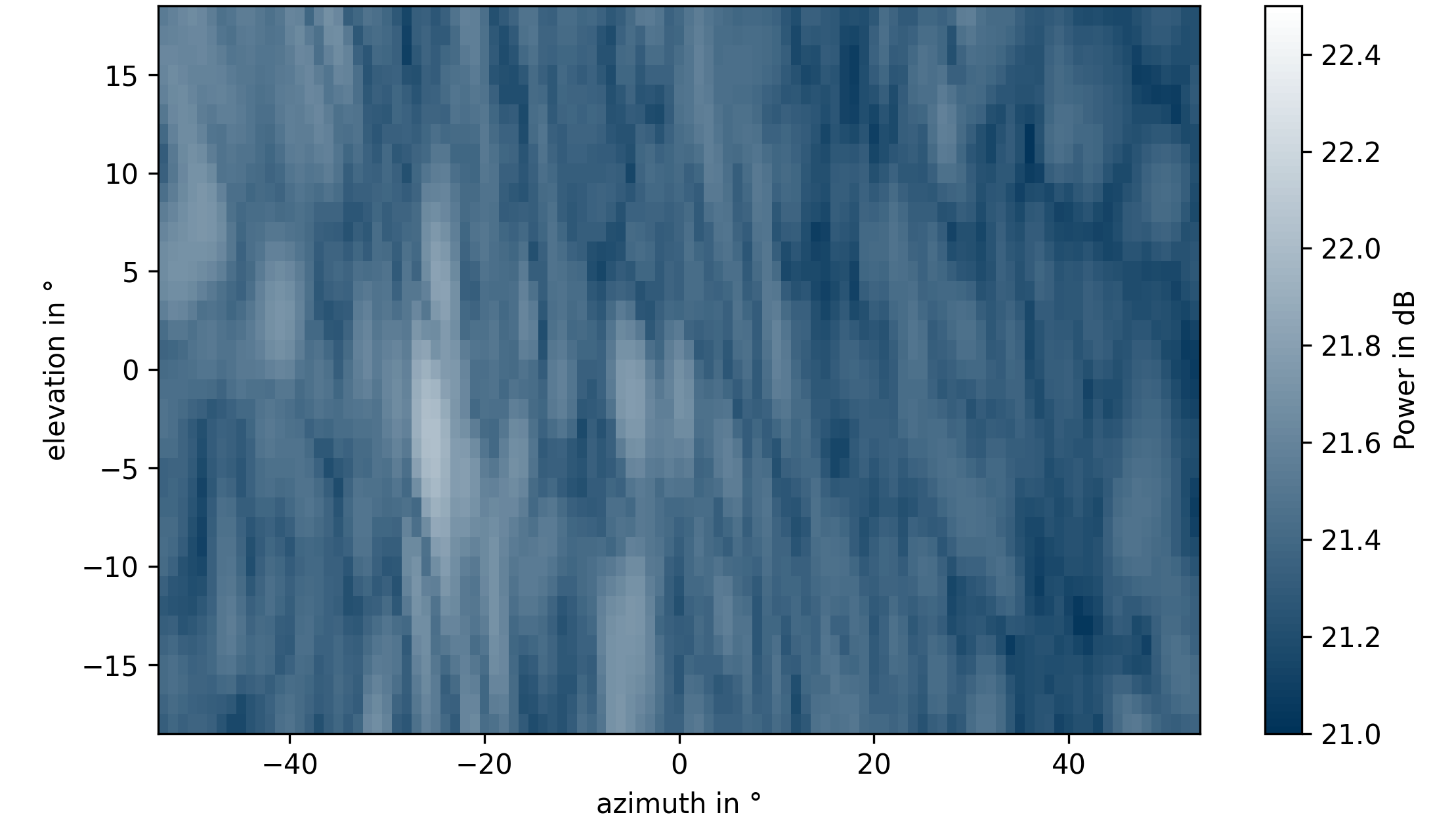}
    }
    \hfill
    \subfloat[\empty]{%
        \includegraphics[width=0.32\linewidth]{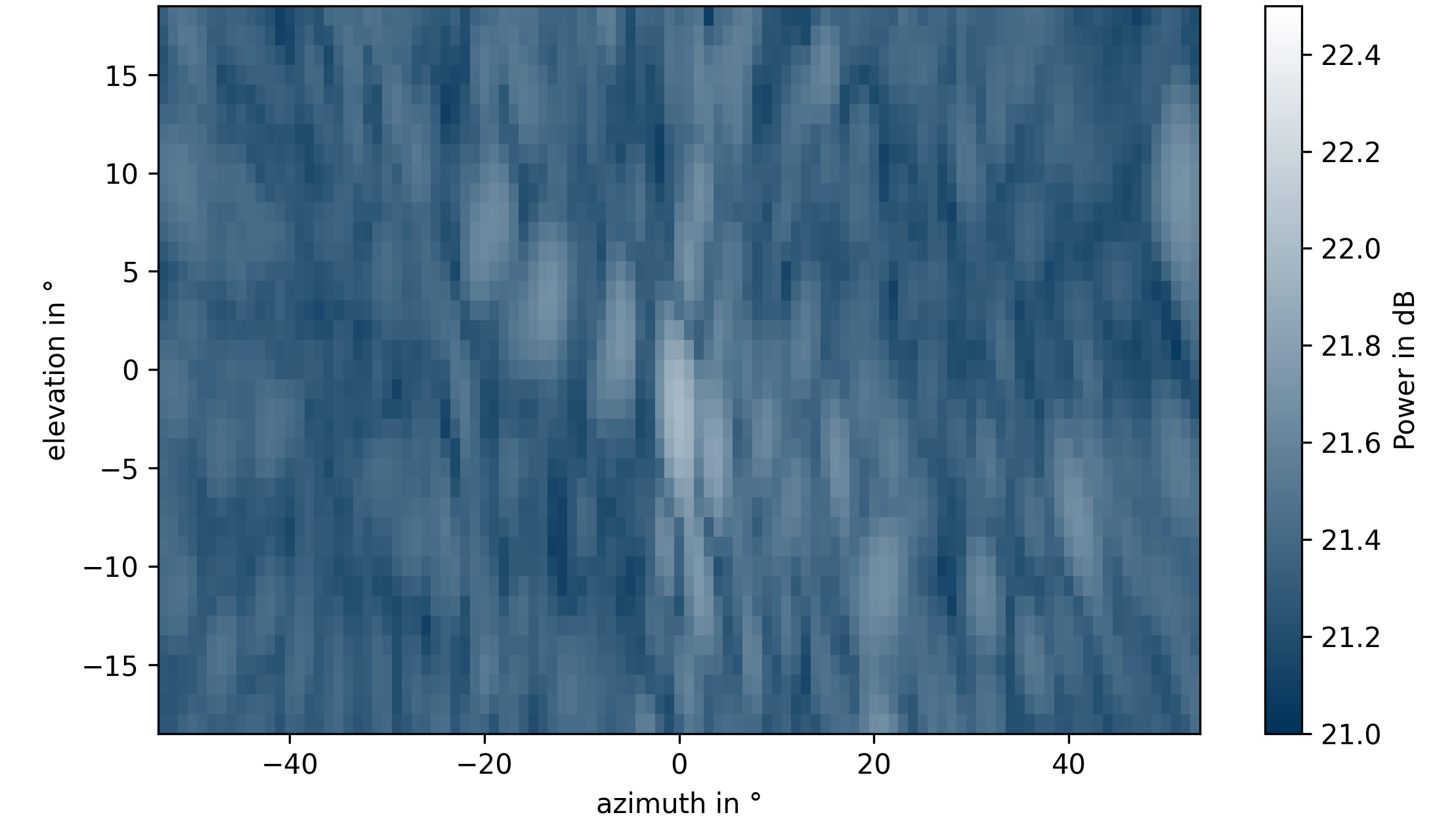}
    }
    \hfill
    \subfloat[\empty]{%
        \includegraphics[width=0.32\linewidth]{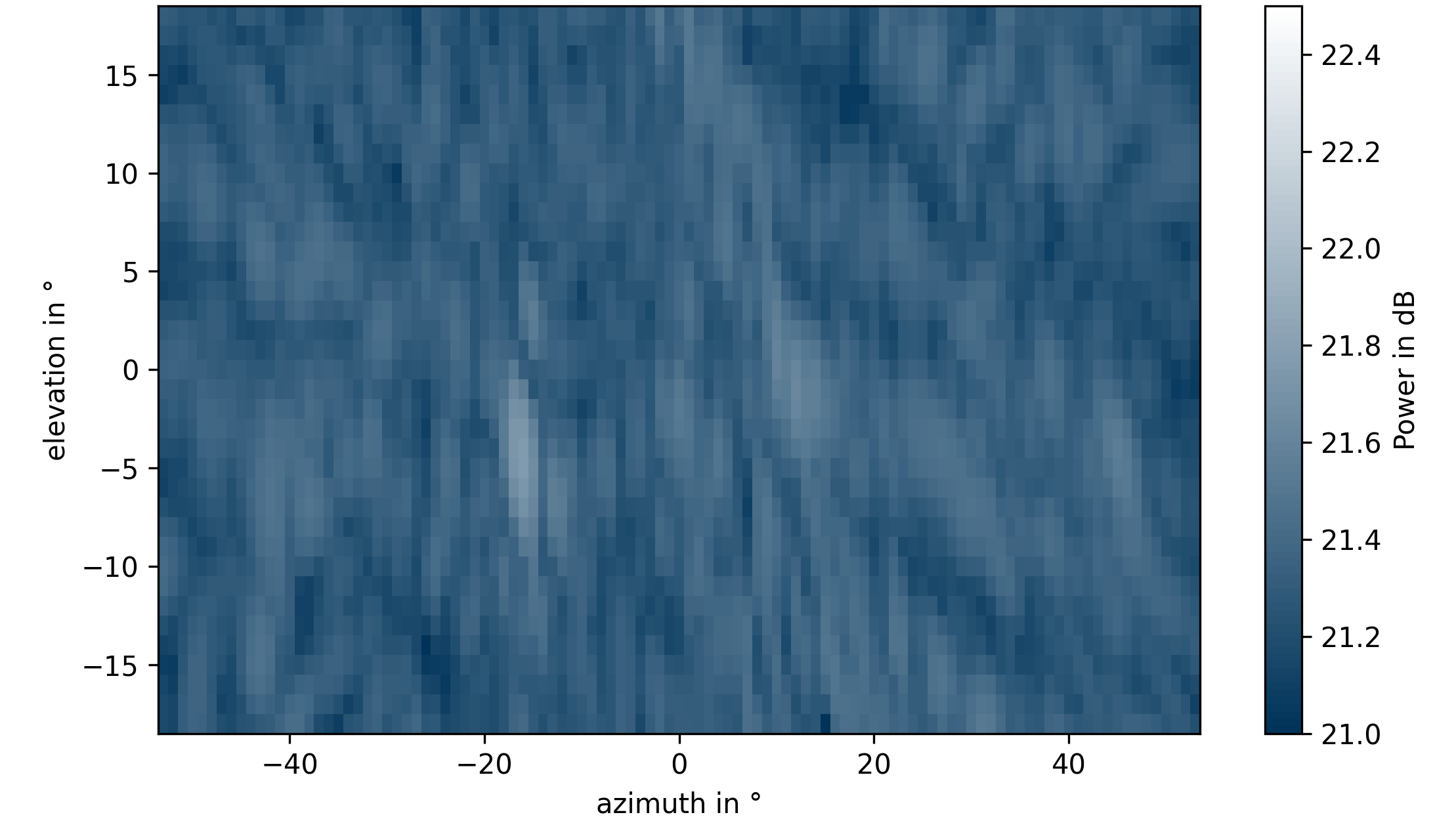}
    }
    \\ \vspace{-4pt}
    \subfloat[\empty]{%
        \includegraphics[width=0.25\linewidth]{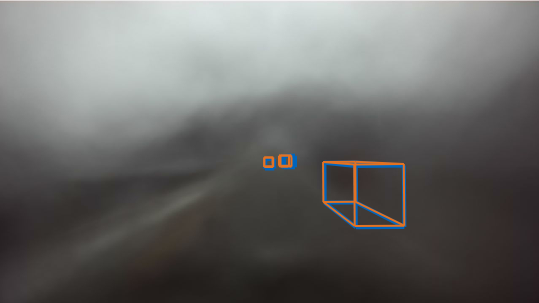}
    }
    \hspace{38pt}
    \subfloat[\empty]{%
        \includegraphics[width=0.25\linewidth]{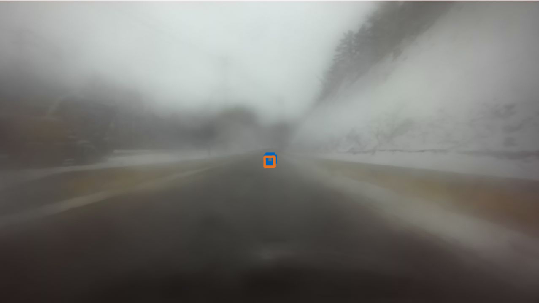}
    }
    \hspace{38pt}
    \subfloat[\empty]{%
        \includegraphics[width=0.25\linewidth]{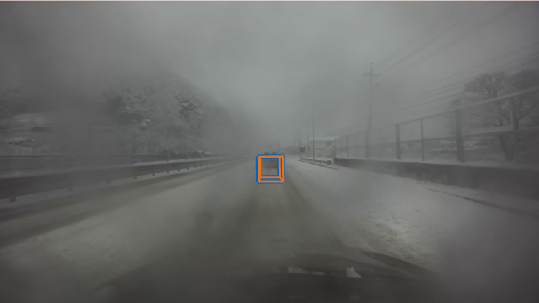}
    }
    \setcounter{subfigure}{3}
    \\ \vspace{-4pt}
    \subfloat[droplet]{%
        \includegraphics[width=0.32\linewidth]{appendix/3_00252_00222_night_ra.png}
    }
    \hfill
    \subfloat[fog]{%
        \includegraphics[width=0.32\linewidth]{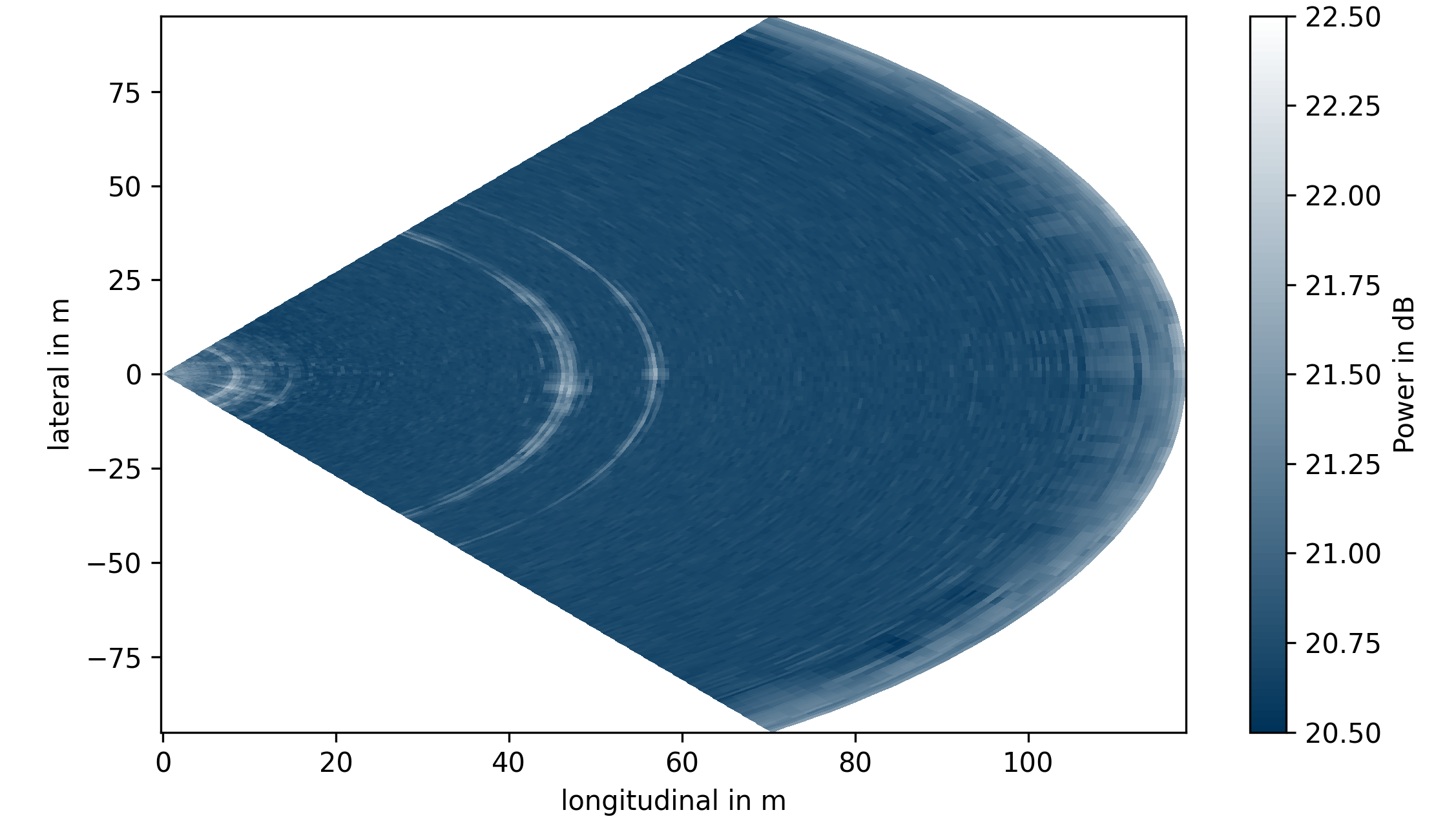}
    }
    \hfill
    \subfloat[snow]{%
        \includegraphics[width=0.32\linewidth]{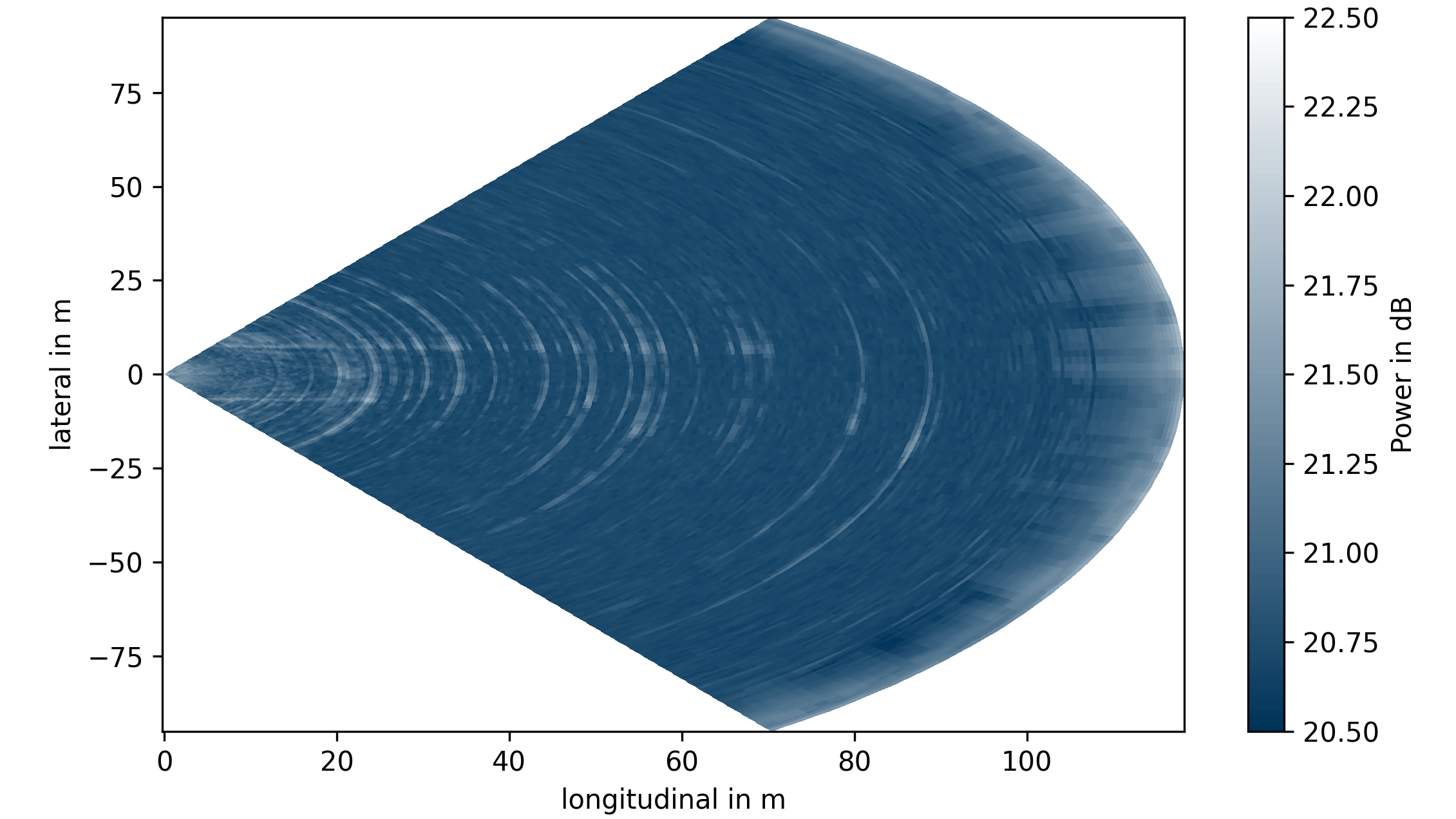}
    }
    
    \caption{Exemplary results of the model performance under night, rain, and snow conditions. The camera data is shown in the center, the radar range-azimuth (RA) data at the bottom, and the radar azimuth-elevation (AE) data at the top. The ground truth is shown in \textcolor{tumblue}{blue} and the model prediction in \textcolor{tumorange}{orange}.}
    \label{fig:appx_examples}
\end{figure*}

Figure~\ref{fig:appx_examples} shows the model predictions and ground truth data plotted onto the camera images and the associated radar data in the range-azimuth (RA) and azimuth-elevation (AE) planes under different environmental conditions.

\bigbreak
\bigbreak
\bigbreak
\null

